\documentclass[a4paper,11pt]{article}

\usepackage[colorlinks = true,
            linkcolor = magenta,
            urlcolor  = blue,
            citecolor = green,
            anchorcolor = blue]{hyperref}
\usepackage{url}
\usepackage[blocks]{authblk}  % authors with affiliations
\usepackage[utf8]{inputenc} % allow utf-8 input
\usepackage[T1]{fontenc}    % use 8-bit T1 fonts
\usepackage{url}            % simple URL typesetting
\usepackage{booktabs}       % professional-quality tables
\usepackage{amsfonts}       % blackboard math symbols
\usepackage{nicefrac}       % compact symbols for 1/2, etc.
\usepackage{enumitem}
\usepackage{subcaption}
\usepackage{float} % Add this in the preamble if not already included
\usepackage[top=2.5cm, bottom=2.5cm, left=2.5cm, right=2.5cm]{geometry}
\usepackage{qtree}
\usepackage{svg}            % For including svg
\usepackage[linguistics]{forest}
\usepackage{amsmath,amsfonts,bm}

\def\eqref#1{equation~\ref{#1}}
\def\1{\bm{1}}

\DeclareMathAlphabet{\mathsfit}{\encodingdefault}{\sfdefault}{m}{sl}
\SetMathAlphabet{\mathsfit}{bold}{\encodingdefault}{\sfdefault}{bx}{n}

\newcommand{\din}{D}
\newcommand{\dactive}{S}
\newcommand{\noise}{\sigma_\text{flip}}

\newcommand{\demb}{D_\text{emb}}

\newcommand{\dfit}{\Omega_{\mathrm{fit}}}

\usepackage{microtype}      % microtypography
\usepackage{graphicx}

\usepackage{amsmath,amssymb,amsthm,bm,dsfont,mathtools}
\usepackage{array, booktabs, makecell}
\newcolumntype{L}{>{$}l<{$}} % math-mode version of "l" column type
\usepackage{multirow}
\usepackage[sorting=none,doi=false,isbn=false,giveninits=true]{biblatex}           % modern bibliography

\usepackage[capitalize,noabbrev]{cleveref}
\newcommand{\citet}[1]{\textcite{#1}}
\newcommand{\citep}[1]{\parencite{#1}}

\bibliography{biblio}

\title{Boolformer: Symbolic Regression of Logic Functions with Transformers}

\author[1]{Stéphane d'Ascoli\thanks{Equal contributions.}}
\author[2]{Arthur Renard \protect\footnotemark[1]\thanks{Corresponding author: arthurenard@icloud.com}}
\author[2]{Vassilis Papadopoulos}
\author[3]{Josh Susskind}
\author[3]{Samy Bengio}
\author[3,4]{Emmanuel Abbé}
\affil[ ]{\textsuperscript{1}EPFL; \textsuperscript{2}CSFT, EPFL; \textsuperscript{3}Apple; \textsuperscript{4};SB/IC, EPFL}

\date{}
\begin{document}
\maketitle
\vspace{-1cm}
\begin{abstract}
We introduce Boolformer, a Transformer-based model trained to perform end-to-end symbolic regression of Boolean functions. First, we show that it can predict compact formulas for complex functions not seen during training, given their full truth table. Then, we demonstrate that even with incomplete or noisy observations, Boolformer is still able to find good approximate expressions. We evaluate Boolformer on a broad set of real-world binary classification datasets, demonstrating its potential as an interpretable alternative to classic machine learning methods. Finally, we apply it to the widespread task of modeling the dynamics of gene regulatory networks and show through a benchmark that Boolformer is competitive with state-of-the-art genetic algorithms, with a speedup of several orders of magnitude. Our code and models are available publicly.
\end{abstract}

\newcommand{\com}[1]{\textcolor{red}{#1}}

\section{Introduction}
\label{sec:introduction}

Deep neural networks, in particular those based on the Transformer architecture~\citep{vaswani2017attention}, have led to breakthroughs in computer vision~\citep{dosovitskiy2020image} and language modelling~\citep{brown2020gpt}, and have fuelled the hopes to accelerate scientific discovery~\citep{jumper2021highly}. 
However, their ability to perform simple logic tasks remains limited~\citep{deletang2022neural}. These tasks differ from traditional vision or language tasks in the combinatorial nature of their input space, which makes representative data sampling challenging. 
%For example, any training set will have a maximal length for its inputs, and one can always ask for the prediction on inputs having twice that length. It is thus crucial to ensure that the model has the right algorithmic bias to hope for out-of-distribution generalization. 

Reasoning tasks have thus gained major attention in the deep learning community, either (i) with explicit reasoning in the logical domain, e.g., tasks in the realm of arithmetic~\citep{saxton2019analysing,lewkowycz2022solving}, algebra~\citep{zhang2022unveiling} or 
algorithmics~\citep{velivckovic2022clrs}, or (ii) implicit reasoning in other modalities, e.g., benchmarks such as Pointer Value Retrieval~\citep{zhang2021pointer} and Clevr~\citep{johnson2017clevr} for vision models, or LogiQA~\citep{liu2020logiqa} and GSM8K~\citep{cobbe2021training} for language models. Reasoning also plays a key role in tasks which can be tackled via Boolean modelling, particularly in the fields of biology~\citep{wang2012boolean} and medicine~\citep{hemedan2022boolean}.

As these endeavours remain challenging for current Transformer architectures, it is natural to examine whether they can be handled more effectively with different approaches, e.g., by better exploiting the Boolean nature of the task.
In particular, when learning Boolean functions with a `classic' approach based on minimizing the training loss on the outputs of the function, Transformers learn potentially complex interpolators as they focus on minimizing the degree profile in the Fourier spectrum, which is not the type of bias desirable for generalization on domains that are not well sampled~\citep{abbe2022learning}. In turn, the complexity of the learned function makes its interpretability challenging. This raises the question of how to improve generalization and interpretability of such models.

\begin{figure*}[tb]
    \centering
    \begin{minipage}{0.47\textwidth}
        \centering
        \large
        \begin{forest}
[$\mathrm{or}$ [$\substack{\text{gill size}}$ ][$\substack{\text{ring}\\\text{type=3}}$ ][$\mathrm{and}$ [$\substack{\text{stalk}\\\text{root=1}}$ ][$\substack{\text{cap}\\\text{ surface=3}}$ ]][$\mathrm{and}$ [$\substack{\text{stalk surface} \\ \text{below ring=2}}$ ][$\mathrm{or}$ [$\substack{\text{stalk}\\\text{root=1}}$ ][$\substack{\text{gill size}}$ ]]]]
            \end{forest}
            \caption{\textbf{A Boolean formula predicted to determine whether a mushroom is poisonous.} We considered the "mushroom" dataset from the PMLB database~\citep{Olson2017PMLB}, and this formula achieves an F1 score of 0.96.}
            \label{fig:mushroom}
    \end{minipage}
    \hfill
    \begin{minipage}{0.37\textwidth}
        \centering
        \includegraphics[width=\linewidth]{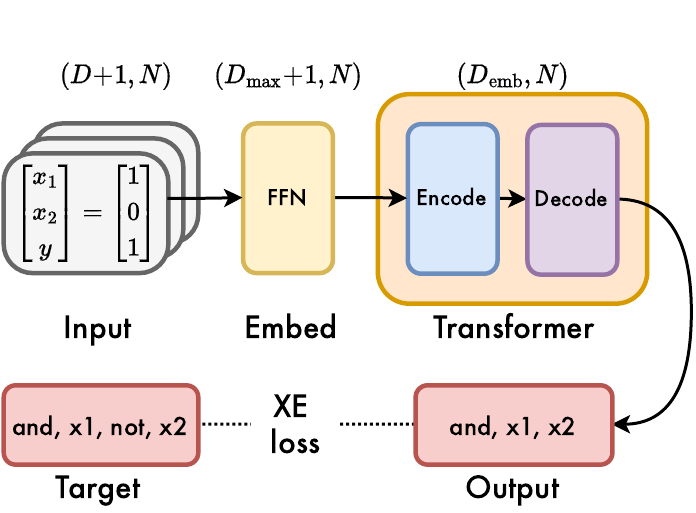}
        \caption{\textbf{Summary of our approach.} We feed $N$ points $(\boldsymbol x,f(\boldsymbol x))\in\{0,1\}^{\din+1}$ to a seq2seq Transformer, and supervise the prediction to $f$ via cross-entropy loss.}
        \label{fig:method}
    \end{minipage}
\end{figure*}

In this paper, we tackle Boolean function learning with Transformers, but we rely directly on `symbolic regression': our Boolformer is tasked to directly predict a Boolean formula, i.e., a symbolic expression of the Boolean function in terms of the three fundamental logical gates (AND, OR, NOT) such as those of Figs.~\ref{fig:multiplexer},\ref{fig:mushroom}. As illustrated in Fig.~\ref{fig:method}, this task is framed as a sequence prediction problem: each training example is a synthetically generated function whose truth table is the input and whose formula is the target.
%When letting the model explore Boolean functions in the free overparametrized setting, fairly complex solutions may be produced that behave well in-distribution but widely out-distribution, even when targets may have simple interpretations such as for the majority example detiled in []. 

By moving to this setting, we decouple the symbolic task of inferring the logical formula and the numerical task of evaluating it on new inputs: the Boolformer only has to handle the first part. We show that this approach can give surprisingly strong performance both in abstract and real-world settings, and discuss how this lays the ground for future improvements and applications.

\begin{figure}[tb]
    \footnotesize
    % \begin{subfigure}[b]{.20\textwidth}            
    %     \footnotesize
    %     \centering
        \begin{forest}
        [$\mathrm{or}$ [$\mathrm{and}$ [$s_0$ ][$\mathrm{or}$ [$s_1$ ][$x_1$ ]][$\mathrm{or}$ [$\mathrm{not}$ [$s_1$ ]][$x_3$ ]]][$\mathrm{and}$ [$\mathrm{not}$ [$s_0$ ]][$\mathrm{or}$ [$s_1$ ][$x_0$ ]][$\mathrm{or}$ [$\mathrm{not}$ [$s_1$ ]][$x_2$ ]]]]
        \end{forest}
        % \caption{Multiplexer}
    % \end{subfigure}
    % \hfill
    % \begin{subfigure}[b]{.70\textwidth}            
        \centering
    % \makebox[0.7\textwidth]{
    %     \begin{forest}
    %     [$\mathrm{and}$ [$\mathrm{or}$ [$x_0$ ][$\mathrm{not}$ [$x_5$ ]]][$\mathrm{or}$ [$\mathrm{and}$ [$x_0$ ][$\mathrm{not}$ [$x_5$ ]]][$\mathrm{and}$ [$x_1$ ][$\mathrm{not}$ [$x_6$ ]]][$\mathrm{and}$ [$\mathrm{or}$ [$x_1$ ][$\mathrm{not}$ [$x_6$ ]]][$\mathrm{or}$ [$x_2$ ][$\mathrm{not}$ [$x_7$ ]]][$\mathrm{or}$ [$\mathrm{and}$ [$x_2$ ][$\mathrm{not}$ [$x_7$ ]]][$\mathrm{and}$ [$x_3$ ][$\mathrm{not}$ [$x_8$ ]]][$\mathrm{and}$ [$x_4$ ][$\mathrm{not}$ [$x_9$ ]][$\mathrm{or}$ [$x_3$ ][$\mathrm{not}$ [$x_8$ ]]]]]]]]
    %     \end{forest}
    %     }
        % \caption{Comparator}
    % \end{subfigure}
    \caption{\textbf{Example Boolean formula predicted by our model.} Depicted: the multiplexer, a function commonly used in electronics to select one out of four sources $x_0, x_1, x_2, x_3$ based on two selector bits $s_0, s_1$.}
    \label{fig:multiplexer}
\end{figure}

\subsection{Contributions}

\begin{enumerate}[leftmargin=*,noitemsep]
\item We train Transformers over synthetic datasets to perform end-to-end symbolic regression of Boolean formulas. The synthetic functions are generated by simplifying formulas whose tree have either low width or low depth (see Section \ref{genf}). We show that given the full truth table of an unseen function, Boolformer is able to predict a compact formula, as illustrated in Fig.~\ref{fig:multiplexer}.

\item We show that Boolformer is robust to noisy and incomplete observations, by training it on incomplete truth tables with flipped bits and irrelevant variables. This is a necessary condition for its applicability to real-word data.

\item We evaluate Boolformer on various real-world binary classification tasks from the PMLB database~\citep{Olson2017PMLB} and show that it is competitive with classic machine learning approaches such as Random Forests while being more interpretable as illustrated in Fig.~\ref{fig:mushroom}.

\item We apply Boolformer to the well-studied task of modeling Gene Regulatory Networks (GRNs) in biology. Using a recent benchmark ~\cite{puvsnik2022review}, we show that our model is competitive with state-of-the-art methods with orders of magnitude faster inference. 
\end{enumerate}

\paragraph{Reproducibility}

All code used for the paper is available at \url{https://github.com/arthurenard/Boolformer}.

\subsection{Related Work}
\label{related}

\paragraph{Logical reasoning in deep learning}

Several papers have studied the ability of deep neural networks to solve logic tasks. \citet{evans2018learning} introduce differential inductive logic as a method to learn logical rules from noisy data, and a few subsequent works attempted to craft dedicated neural architectures to improve this ability~\citep{ciravegna2023logic,shi2020neural,dong2019neural}.
Large language models (LLMs) such as ChatGPT, however, have been shown to perform poorly at simple logical tasks such as basic arithmetic~\citep{deletang2022neural,jelassi2023length}, and tend to rely on approximations and shortcuts~\citep{liu2022transformers}. Although some reasoning abilities seem to emerge with scale~\citep{wei2022emergent} and can be enhanced via several procedures such as scratchpads~\citep{nye2021show} and chain-of-thought prompting~\citep{wei2022chain}, achieving holistic and interpretable reasoning in LLMs remains a challenge.

\paragraph{Learning Boolean functions}

Learning Boolean functions has been an active area in theoretical machine learning, mostly under the probably approximately correct (PAC) and statistical query (SQ) learning frameworks~\citep{hellerstein2007pac,reyzin2020statistical}. 
More recently, \citet{abbe2023sgd} shows that regular neural networks learn by gradually fitting monomials of increasing degree, in such a way that the sample complexity is governed by the  `leap complexity' of the target function, i.e. the largest degree jump the Boolean function sees in its Fourier decomposition.
In turn, \citet{abbe2022learning} shows that this leads to a `min-degree bias' limitation: Transformers tend to learn interpolators having least `degree profile' in the Boolean Fourier basis, which typically lose the Boolean nature of the target and often produce complex solutions with poor out-of-distribution generalization. 

\paragraph{Inferring Boolean formulas}

A few works have explored the paradigm of inferring Boolean formulas in symbolic form, using SAT solvers~\citep{narodytska2018learning}, ILP solvers~\citep{wang2015learning,su2015interpretable} or  LP-relaxation~\citep{malioutov2017learning}. However, all these works predict the formulas in conjunctive or disjunctive normal forms (CNF/DNF), which typically amounts to exponentially long formulas. In contrast, the Boolformer is biased towards predicting compact expressions\footnote{Consider for example the comparator of Fig.~\ref{fig:multiplexer}: since the truth table has roughly as many positive and negative outputs, the CNF/DNF involves $\mathcal{O}(2^D)$ terms where $D$ is the number of input variables. For $D=10$ this is several thousand binary gates, versus 17 for our model.}, which is more akin to logic synthesis -- the task of finding the shortest circuit to express a given function, also known as the Minimum Circuit Size Problem (MCSP). While a few heuristics (e.g. Karnaugh maps~\citep{karnaugh1953map}) and algorithms (e.g. ESPRESSO~\citep{rudell1987espresso}) exist to tackle the MCSP, its NP-hardness~\citep{murray2017mscp} remains a barrier towards efficient circuit design. Given the high resilience of computers to errors, approximate logic synthesis techniques have been introduced~\citep{scarabottolo2018circuit,venkataramani2012salsa,venkataramani2013substitute,boroumand2021learning,oliveira1993learning,rosenberg2023explainable}, with the aim of providing approximate expressions given incomplete data -- this is similar in spirit to what we study in Section~\ref{sec:noisy}.

\paragraph{Symbolic regression}

Symbolic regression (SR), i.e. the search of mathematical expressions underlying a set of numerical values, is still today a rather unexplored paradigm in the ML literature. Since this search cannot directly be framed as a differentiable problem, the dominant approach for SR is genetic programming
%~\citep{la2018learning,mcconaghy2011ffx,virgolin2021improving,de2021interaction,arnaldo2014multiple,virgolin2019linear,kommenda2020genetic} 
(see~\citep{la2021contemporary} for a recent review). 
A few recent publications applied Transformer-based approaches to SR~\citep{biggio2021neural,valipour2021symbolicgpt,kamienny2022end,tenachi2023deep}, yielding comparable results but with a significant advantage: the inference time rarely exceeds a few seconds, several orders of magnitude faster than existing methods. Indeed, while the latter needs to be run from scratch on each new set of observations, Transformers are trained over synthetic datasets, and inference simply consists of a forward pass. Such efficiency opens the possibility of merging the two approaches, by using the trained model to suggest meaningful mutations in the context of genetic algorithm (e.g. \cite{romera-paredes_mathematical_2024}). This may boost both the convergence speed and the quality of the results, especially in search spaces where most random mutations would result in invalid candidates.

\section{Methods} 
\label{sec:methods}
Our task is to infer Boolean functions of the form $f: \{0, 1\}^D \to \{0,1\}$, by predicting a Boolean formula built from the basic logical operators: AND, OR, NOT, as illustrated in Figs.~\ref{fig:multiplexer},\ref{fig:mushroom}. We train Transformers~\citep{vaswani2017attention} on a large dataset of synthetic examples, following the seminal approach of~\cite{lample2019deep}. For each example, the input $\dfit$ is a set of pairs $\{(\boldsymbol x_i, y=f(\boldsymbol{x}_i))\}_{i=1\ldots N}$, and the target is the function $f$ as described above. Our general method is summarized in Fig.~\ref{fig:method}. Examples are generated by first sampling a random function $f$, then generating the corresponding $(\boldsymbol{x}, y)$ pairs as described in the following sections.

\subsection{Generating formulas}\label{genf}
To sample Boolean formulas\footnote{A Boolean formula is a tree where input variables can appear more than once, and differs from a Boolean circuit, which is a directed graph which can feature cycles, but where each input bit appears once at most.}, we construct random unary-binary tree with mathematical operators at the internal nodes and variables at the leaves. We rely on the tree generator of \cite{lample2019deep}, whose distribution is biased towards trees which are either relatively narrow (and possibly deep) or relatively shallow (and possibly wide). Moreover, once operators and variables are sampled inside the tree, we further simplify the formula using algebraic rules in order to make the formula as simple as possible, encouraging the model to predict maximally compact formulas. The full sampling procedure is detailed in App.~\ref{app:generator}.

The distribution of functions generated in this way spans the whole space of possible Boolean functions (of size $2^{2^\din}$), but in a non-uniform fashion with a bias towards functions described by a relatively simple formula\footnote{Indeed, for uniformly random function, the task would be hopeless as it is known to be NP-hard~\citep{murray2017mscp}}. As discussed quantitatively in App.~\ref{app:memorization}, the diversity of functions generated in this way is such that throughout the whole training procedure, functions of dimension $\din\geq 7$ are typically encountered at most once. 

%\footnote{More involved generation procedures, e.g. involving Boolean circuits, could be envisioned as discussed in Sec.~\ref{sec:discussion}, but we leave this for future work.}

\subsection{Generating inputs}
\label{sec:inputs}

% We then assess the effective support of the function $\tilde D$, i.e. the number of input variables which affect the output. To do so, we sample 1000 points in the hypercube, and check whether their image is changed when flipping each input bit. If $\tilde D < D/2$, we reject the function: in other words, we require at least half of the input variables to be active.

% $\tilde D$ is a good indicator of the sample complexity, as the number of examples required to fully determine a $\tilde{D}$-dimensional function is $N_\star = 2^{\tilde D}$. Hence, we use $\tilde D$ to determine the number of input samples to feed to the model:
% $N \sim \mathcal U (N_\star/4, N_\star)$.

Once the function $f$ is generated, we select $N$ points $\boldsymbol x$ in the Boolean hypercube following the procedure detailed below, and compute the corresponding outputs $y=f(\boldsymbol x)$. Optionally, we may flip the bits of the inputs and outputs independently with probability $\noise$; we consider the two following setups.

\paragraph{Noiseless regime}
The noiseless regime, studied in Sec.~\ref{sec:noiseless}, is defined as follows:
\begin{itemize}[leftmargin=*,noitemsep]
    \item \textbf{Noiseless data:} there is no bit flipping, i.e. $\noise=0$.
    \item \textbf{Full support:} all the input variables are present in the Boolean formula.
    \item \textbf{Full observability:} the model has access to the whole truth table of the Boolean function, i.e. $N = 2^\din$. This limits us to a maximum of 10 input variables.
\end{itemize} 
\paragraph{Noisy regime}
In the noisy regime, studied in Sec.~\ref{sec:noisy}, the model must determine which variables affect the output, while also being able to cope with corruption of the inputs and outputs. During training, we vary the amount of noise for each sample so that the model can handle a variety of noise levels:
\begin{itemize}[leftmargin=*,noitemsep]
    \item \textbf{Noisy data:} the probability of each bit (both input and output) being flipped $\noise$ is sampled uniformly in $[0,0.1]$.
    \item \textbf{Partial support:} the model can handle functions with up to 120 input variables, but only up to 6 of these are ``active'', i.e. appear in the Boolean formula -- all the other variables are ``inactive''.
    \item \textbf{Partial observability:} a subset of the hypercube is observed: the number of input points $N$ is sampled uniformly in $[30,300]$, which is typically much smaller that $2^D$. Additionally, instead of sampling uniformly (which would cause distribution shifts if the inputs are not uniformly distributed at inference), we generate the input points via a random walk in the hypercube. Namely, we sample an initial point $\boldsymbol x_0$ then construct the following points by flipping independently each coordinate with a probability sampled uniformly in $[0.05,0.25]$.
\end{itemize}

\subsection{Model}

\paragraph{Tokenization}
% To represent Boolean formulas as sequences fed to the Transformer, we enumerate the nodes of the trees in prefix order, i.e., direct Polish notation as in~\citep{lample2019deep}: operators and variables are represented as single autonomous tokens, e.g. $[\texttt{AND}, x_1, \texttt{NOT}, x_2]$\footnote{We discard formulas which require more than 200 tokens.}. The inputs are embedded using $\{\texttt{0},\texttt{1}\}$ tokens.

To represent Boolean formulas as sequences processed by the decoder, we enumerate the nodes of the trees in prefix order, i.e., direct Polish notation as in~\citep{lample2019deep}: operators and variables are represented as single autonomous tokens, e.g. $[\texttt{AND}, x_1, \texttt{NOT}, x_2]$. The evaluations fed to the encoder are embedded using $\{\texttt{0},\texttt{1}\}$ tokens. In the noiseless regime, we shrink the input length by providing less than half of the truth table, namely only the entries corresponding to the less frequent output of the boolean function. Using a special token, we indicate whether this value is $0$ or $1$ which effectively provides the information of the full truth table, albeit implicitly. Formulas requiring more than 200 tokens are discarded, as we are limited by the attention size of the decoder.

\paragraph{Token Embeddings}
Our model is provided $N$ input points $(\boldsymbol x,y)$, each of which is represented by $D+1$ tokens of dimension $\demb$, where $D$ is the dimension of $\boldsymbol{x}$. As $D$ and $N$ become large, this would result in very long input sequences ($ND$ tokens) which are suboptimal given the quadratic complexity of Transformers in the input length. To mitigate this, we introduce compressed embeddings to map each input pair $(\boldsymbol x,y)$ to a single embedding, following~\cite{kamienny2022end}. To do so we pad the empty input dimensions to $\din_\text{max}$, enabling our model to handle variable input dimension, then concatenate all the tokens and feed the 
$(D_\text{max}+1)\times\demb$-dimensional result into a linear layer which projects it down to dimension $\demb$. The resulting $N$ embeddings of dimension $\demb$ are then fed to the Transformer. 

\paragraph{Transformer}
We use the Transformer architecture~\citep{vaswani2017attention} where the encoder and decoder use 8 and 8 layers, 16 attention heads and an embedding dimension of 512, for a total of  $\sim$ 60M parameters. A notable property of this task is the permutation invariance of the $N$ input points. As such, we remove positional embeddings from the encoder, encoding this invariance in the model. The decoder uses standard learnable, absolute positional embeddings. 

We have attempted scaling up the model, testing 110M and 450M parameter versions. Interestingly, there were few if any improvements, both on the cross-entropy loss, and the benchmark evaluations. This is peculiar, as transformers have shown to yield predictable improvements with scaling \cite{kaplanscale}. We leave to future work the investigation of this phenomenon and its potential relation to the specific task of symbolic regression.

% \begin{figure}[tb]
%     \centering
%     \includegraphics[width=\columnwidth]{figs/attention_1sin_20x_0_x_0-1.png}
%     \caption{\textbf{Attention heads reveal intricate mathematical analysis.} We considered the expression $f(x)=\sin(x)/x$, with $N=100$ input points sampled between $-20$ and $20$ (red dots; the y-axis is arbitrary).  We plotted the attention maps of a few heads of the encoder, which are $N\times N$ matrices where the element $(i,j)$ represents the attention between point $i$ and point $j$. Notice that heads 2, 3 and 4 of the second layer analyze the periodicity of the function in a Fourier-like manner.}
%     \label{fig:attention-small}
% \end{figure}

\subsection{Training and evaluation}

\paragraph{Training}

We optimize a cross-entropy loss with the AdamW optimizer and a batch size of 1024, warming up the learning rate from $10^{-7}$ to $2\times 10^{-4}$ over the first 5,000 steps. It is then kept constant for 60,000 steps, after which we perform a linear cooldown back to $0$ \cite{hagele2024scalinglawscomputeoptimaltraining}. On 4 H100 GPUs, this takes about 1 day.

\paragraph{Inference}
At inference time, we find that beam search does not provide measurable improvements compared to standard sampling. Therefore, in most results presented in this paper, we generate 10 candidates by sampling, then rank them according to how well they fit the input data (to assess this, we use the fitting error defined below). Note that when the data is noiseless, the model will often find several candidates which have an accuracy close to $100\%$.
\paragraph{Evaluation}
Given a set of input-output pairs $\Omega$ generated by a target function $f_\star$, we compute the accuracy of a predicted function $f$ as $
\frac{1}{|\Omega|}\sum_{(\boldsymbol x, y) \in \Omega} 1[f(\boldsymbol x) = f_\star(\boldsymbol{x})]$.

We can then define:
\begin{itemize}[leftmargin=*, noitemsep]
    \item \textbf{Fitting accuracy:} accuracy obtained when re-using the points used to predict the formula, $\Omega = \dfit$
    \item \textbf{Fitting perfect recovery:} defined as 1 if the fitting error is strictly equal to 0, and 0 otherwise.
    \item \textbf{Test accuracy:} accuracy obtained when sampling points different than the ones used for the prediction. Thie metric does not apply in the noiseless regime, as the model observes all possible sampling points. 
    \item \textbf{Test perfect recovery:} defined as 1 if the test error is strictly equal to 0, and 0 otherwise.
\end{itemize}

\section{Noiseless regime: finding the shortest formula}
\label{sec:noiseless}
The noiseless setting is akin to logic synthesis, where the goal is to find the shortest formula that implements a given function. The practical relevance of this regime is limited by the fact that it requires full observability of the function; as such, the results of this section serve as a controlled setting to probe the generalization abilities of Boolformer. Specifically, we do not claim that logic synthesis is a useful application of Boolformer in its current iteration, but rather an interesting task to examine the potential of the model, which we argue really shines in the noisy setting (Sec. \ref{sec:noisy}).

\vspace{.2cm}
\noindent
\begin{minipage}{.40\linewidth}
\paragraph{In-domain performance}  
In Fig.~\ref{fig:error-acc-noiseless}, we report the performance of the model when varying the number of input variables and the number of operators in the ground truth. Metrics are averaged over $10,000$ formulas, sampled from the generator used during training.

The model demonstrates high accuracy in predicting target functions across all cases, including for $D \geq 7$, where memorization is not feasible (samples with $D \geq 7$ have typically not been encountered during training  as shown in App.~\ref{app:memorization}).
\end{minipage}
\hfill
\begin{minipage}{.53\linewidth}
    \centering
    \includegraphics[width=\linewidth]{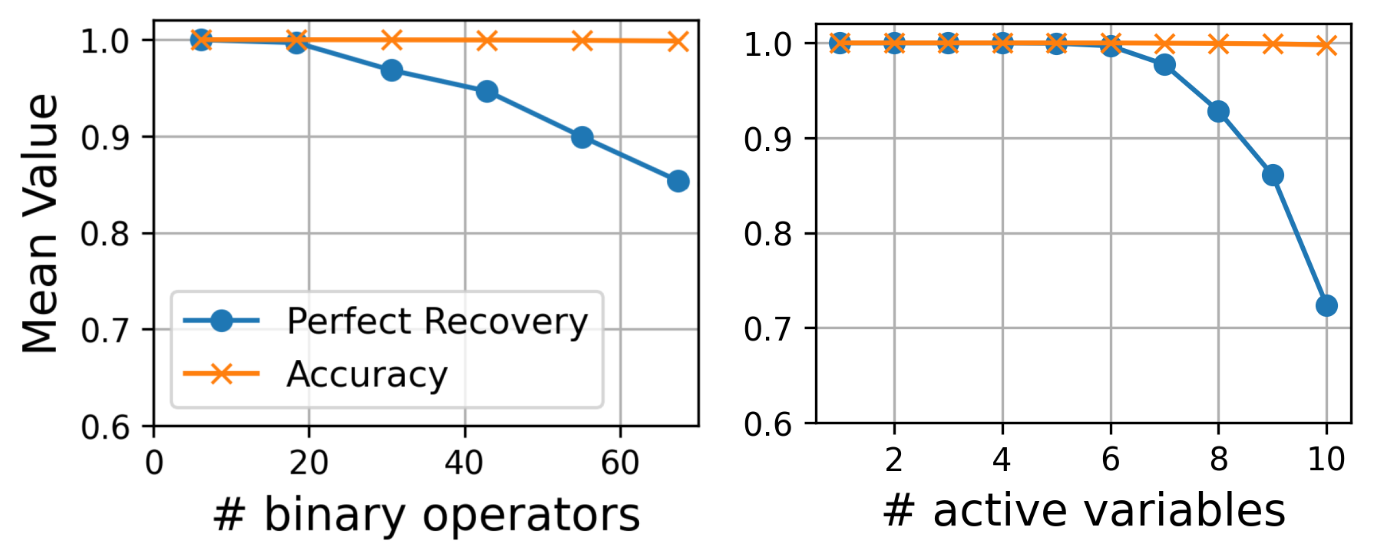}
    % \begin{subfigure}[t]{0.515\linewidth}
    %     \includegraphics[width=\linewidth]{new_figs/binary_op_noiseless.png}
    %     % \caption{Part 1 description.}
    % \end{subfigure}
    % \hfill
    % \begin{subfigure}[t]{0.475\linewidth}
    %     \includegraphics[width=\linewidth]{new_figs/active_var_vs_metrics.png}
    %     % \caption{Part 2 description.}
    % \end{subfigure}
    \captionof{figure}{\textbf{Our model is able to approximate the formula of unseen functions with high accuracy.} We report perfect recovery and fitting accuracy of our model when varying the number of binary gates and input variables. Metrics are averaged over 10,000 samples from the function generator.}
    \label{fig:error-acc-noiseless}
\end{minipage}

\vspace{0.3cm}
Of course, these results reflect the model's performance on the specific distribution of functions it was trained on, which is highly nonuniform within the $2^{2^\din}$-dimensional space of Boolean functions. It is important to note that for a Boolean function sampled uniformly from this space, the likelihood of achieving any meaningful accuracy is effectively negligible, as such random functions are highly unlikely to be expressible in fewer than 200 tokens.
As a baseline, we also compare with ESPRESSO \cite{rudell1987espresso}, which is a heuristic logic synthesizer that, despite its age, is still relevant today. Comparing to Boolformer (more details in App. \ref{app:espressocompare}), the two methods yield comparable average formula lengths, with Boolformer yielding shorter formulas more often, especially when more variables are involved. The comparison is not perfect, as ESPRESSO is constrained to output formulas in sum-of-products form. Still, the ability to generate concise formulas is confirmation that Boolformer is at least learning non-trivial representations that generalize to unseen data.

\paragraph{Success and failure cases}

In Fig.~\ref{fig:multiplexer}, we show two examples of Boolean functions where our model successfully predicts a compact formula for: the 4-to-1 multiplexer (which takes 6 input variables) and the 5-bit comparator (which takes 10 input variables). In App.~\ref{app:examples}, we provide more examples: addition and multiplication, as well as majority and parity functions. By increasing the dimensionality of each problem up to the point of failure, we show that in all cases our model typically predicts exact and compact formulas as long as the function can be expressed with less than 100 binary gates (which is the largest size seen during training, as larger formulas exceed the 200 tokens limit) and fails beyond. Still, even in this cases, the model is still able to approximate the formula well, as the accuracy still remains high.

Hence, the failure point depends on the intrinsic difficulty of the function: for example, Boolformer can predict an exact formula for the comparator function up to $D=10$, but only $D=6$ for multiplication, $D=5$ for majority and $D=4$ for parity as well as typical random functions (whose outputs are independently sampled from $\{0,1\}$). Parity functions are well-known to be the most difficult functions to learn for SQ models due to their leap-complexity, and are also the hardest to learn in our framework because they require the most operators to be expressed (the XOR operator being excluded in this work).
%Note however that at the breaking point, the model does not fail completely: the error remains rather close to zero.
%However, we show that the sample complexity of $k$-parity functions in the SQ framework, well-known to be $\mathcal{O}(D^k)$, can be easily beaten in the noisy regime described below.

% \begin{figure*}
%     \centering
%     \includegraphics[width=1\linewidth]{figs/error_acc_noisy.pdf}
%     \caption{\textbf{Our model is robust to data incompleteness, bit flipping and noisy variables.} We display the test accuracy of our model when varying the four factors of difficulty described in Sec.~\ref{sec:methods}. The colors depict different number of active variables (which appear in the Boolean formula), as shown in the first panel. Metrics are averaged over 10k samples from the random generator.}
%     \label{fig:error-acc-noisy}
% \end{figure*}

\begin{figure*}
    \centering
    \includegraphics[width=\linewidth]{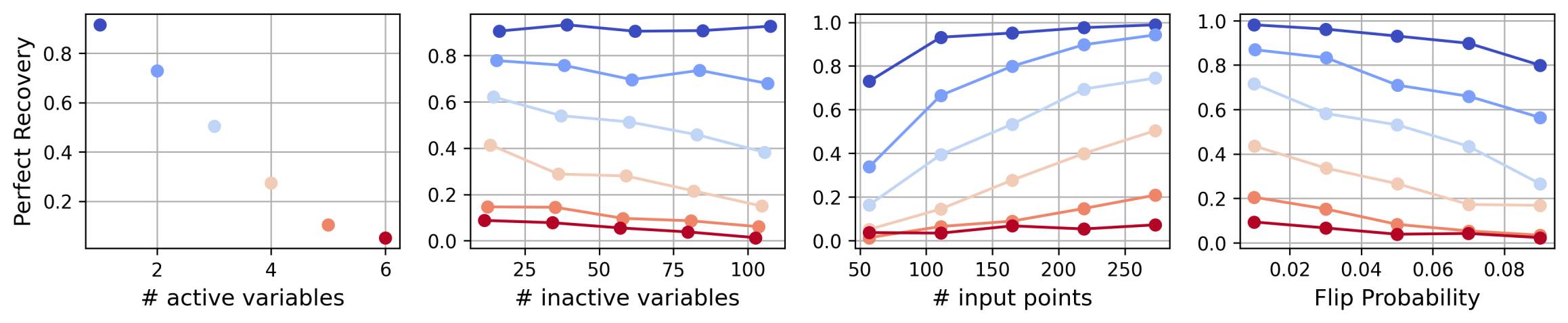}
    % \begin{subfigure}[t]{0.24\linewidth}        \includegraphics[width=\linewidth]{new_figs/perfect_recover_vs_tgt_dim.png}
    %     \label{fig:part1}
    % \end{subfigure}
    % \hfill
    % \begin{subfigure}[t]{0.24\linewidth}
    %     \includegraphics[width=\linewidth]{new_figs/perfect_recover_rdm_vs_nb_inactives.png}
    %     \label{fig:part2}
    % \end{subfigure}
    % \hfill
    % \begin{subfigure}[t]{0.24\linewidth}
    %     \includegraphics[width=\linewidth]{new_figs/perfect_recover_rdm_vs_nb_pts.png}
    %     \label{fig:part3}
    % \end{subfigure}
    % \hfill
    % \begin{subfigure}[t]{0.24\linewidth}
    %     \includegraphics[width=\linewidth]{new_figs/perfect_recover_rdm_vs_prob_flip.png}
    %     \label{fig:part4}
    % \end{subfigure}
 \caption{\textbf{Our model is robust to data incompleteness, bit flipping and noisy variables.} We display the test perfect recovery of our model when varying the four factors of difficulty described in Sec.~\ref{sec:methods}. The colors depict different number of active variables (which appear in the Boolean formula), as shown in the first panel. Metrics are averaged over 10k samples from the random generator.}
    \label{fig:error-acc-noisy}
\end{figure*}

\section{Noisy regime: applications to real-world data}
\label{sec:noisy}

We now turn to the noisy regime, which is defined at the end of Sec.~\ref{sec:inputs}. We begin by studying the robustness of Boolformer to incomplete and corrupted observations, then demonstrate its practical relevance by studying two real-world applications: interpretable binary classification and efficient Gene Regulatory Network (GRN) inference.

\subsection{Results on noisy data}

In Fig.~\ref{fig:error-acc-noisy}, we show how the performance of our model depends on the various factors of difficulty of the problem. The different colors correspond to different numbers of active variables appearing in the formula, as shown in the leftmost panel: in this setting with multiple sources of noise, we see that accuracy drops much faster with the number of active variables than in the noiseless setting.

As could be expected, performance improves as the number of input points $N$ increases, and degrades as the amount of random flipping and the number of inactive variables increase. However, our model copes relatively well with noise in general, as it displays nontrivial generalization even when we add up to 120 inactive variables and up to 10\% random flipping.

\subsection{Application 1: interpretable binary classification}

In this section, we show that our noisy model can be applied to binary classification tasks, providing an interpretable alternative to classic machine learning methods on tabular data.

\paragraph{Method}

We consider the tabular datasets from the Penn Machine Learning Benchmark (PMLB) from~\citep{Olson2017PMLB}. These contain a wide variety of real-world problems, such as predicting chess moves, toxicity of mushrooms, credit scores, and heart diseases. Since our model only takes binary features as input, we discard continuous features and binarize the categorical features with $C>2$ classes into $C$ binary variables. Note that this procedure can greatly increase the total number of features,we only keep datasets for which this results in less than 120 features (the maximum our model can handle). We randomly sample $25\%$ of the examples for testing and report the F1 score obtained on this set.

We compare our model with two classic machine learning methods: logistic regression and random forests, using the default hyperparameters from $\texttt{sklearn}$. For random forests, we test two values for the number of estimators: 1 (in which case we obtain a simple decision tree as for Boolformer) and 100. 

\noindent
\begin{minipage}{.38\linewidth}
\paragraph{Results}
Results are reported in Fig.~\ref{fig:pmlb}, where for readability we only display the datasets where the random forest with 100 estimators achieves an F1 score above 0.75. The performance of Boolformer is similar on average to that of logistic regression : logistic regression typically performs better on "hard" datasets where there is no exact logical rule, for example medical diagnosis tasks such as $\texttt{heart\_h}$, but worse on logic-based datasets where the data is not linearly separable such as $\texttt{xd6}$. 

The F1 score of our model is slightly below that of a random forest of 100 trees, but slightly above that of the random forest with a single tree. This is remarkable considering that the Boolean formula it outputs usually contains a few dozen nodes, whereas the trees of random forest use up to several hundreds. As an example, we display a formula predicted for the mushroom toxicity dataset in Fig.~\ref{fig:mushroom}, and a more extensive collection of formulas in App.~\ref{app:pmlb}.
\end{minipage} \hfill
\begin{minipage}{.6\linewidth}
        \centering
\includegraphics[width=1\linewidth]{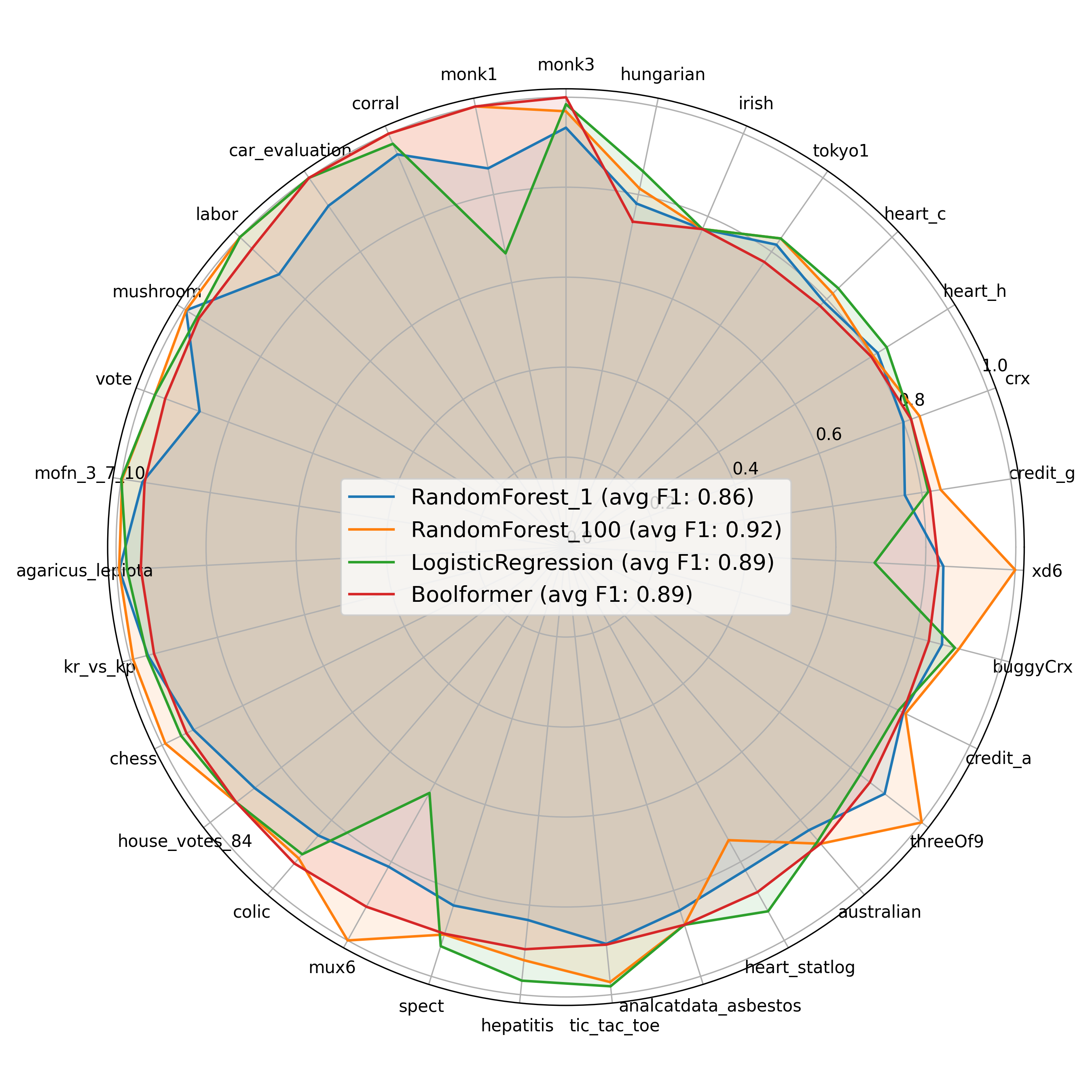}
        \captionof{figure}{\textbf{Our model is competitive with classic machine learning methods while providing highly interpretable results.} We display the F1 score obtained on various binary classification datasets from the Penn Machine Learning Benchmark~\citep{Olson2017PMLB}. We compare the F1 score of the Boolformer with random forests (using 1 and 100 estimators) and logistic regression.}
        \label{fig:pmlb}
\end{minipage}

\subsection{Application 2: efficient inference of gene regulatory networks }
\label{sec:benchmark}

A Boolean network is a dynamical system composed of $D$ bits whose transition from one state to the next is governed by a set of $D$ Boolean functions\footnote{The $i$-th function $f_i$ takes as input the state of the $D$ bits at time $t$ and returns the state of the $i$-th bit at time $t+1$.}. These types of networks have attracted a lot of attention in the field of computational biology as they can be used to model gene regulatory networks (GRNs)~\citep{zhao2021comprehensive} -- see App.~\ref{app:grn} for a brief overview of this field. In this setting, each bit represents the (discretized) expression of a gene (on or off) and each function represents the regulation of a gene by the other genes. In this section, we investigate the applicability of our symbolic regression-based approach to this task. 

\begin{figure*}[tb]
\centering
\begin{subfigure}[b]{0.88\linewidth}
\centering
    \includegraphics[width=\linewidth]{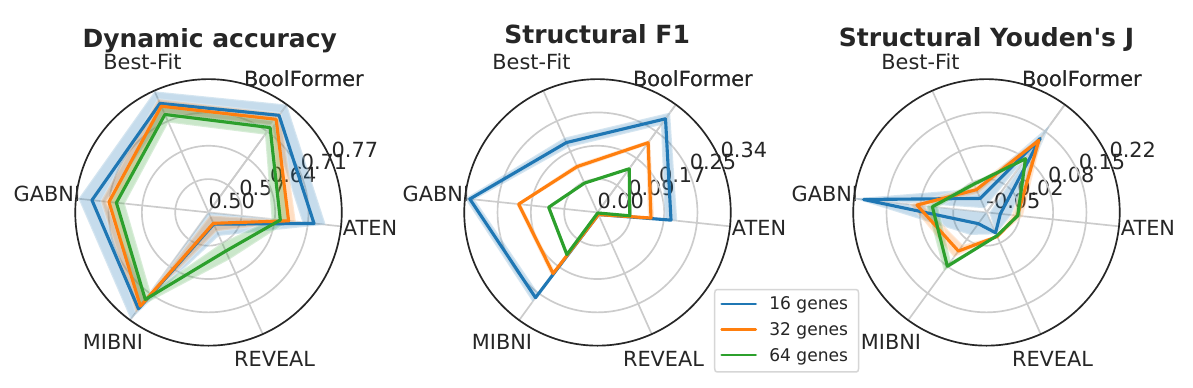}
    \caption{Dynamic and structural metrics}
\end{subfigure}
\begin{subfigure}[b]{.45\linewidth}
    \centering
    \includegraphics[width=\linewidth]{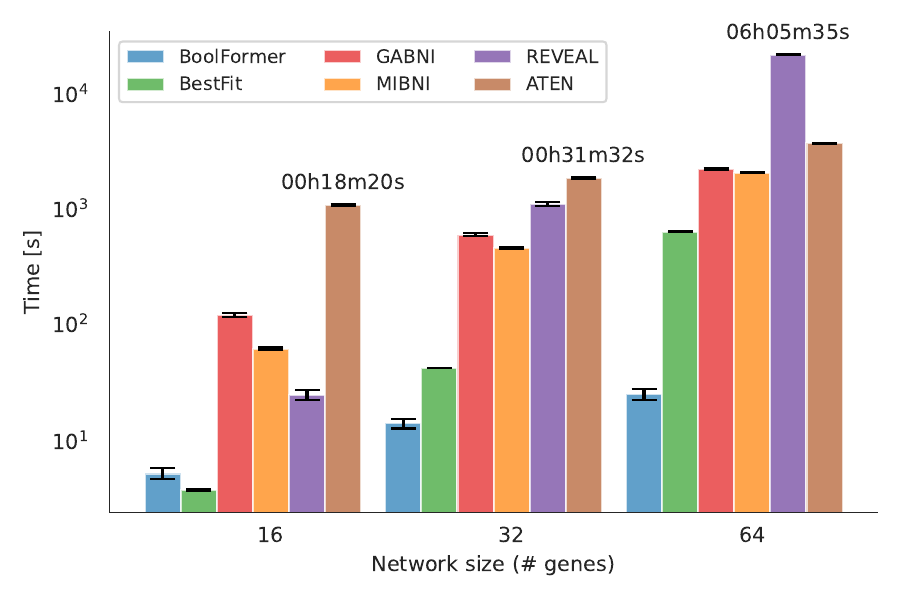}
    \caption{Average inference time}
\end{subfigure}
\hfill
\begin{subfigure}[b]{.4\linewidth}
    \centering
    \includegraphics[width=\linewidth]{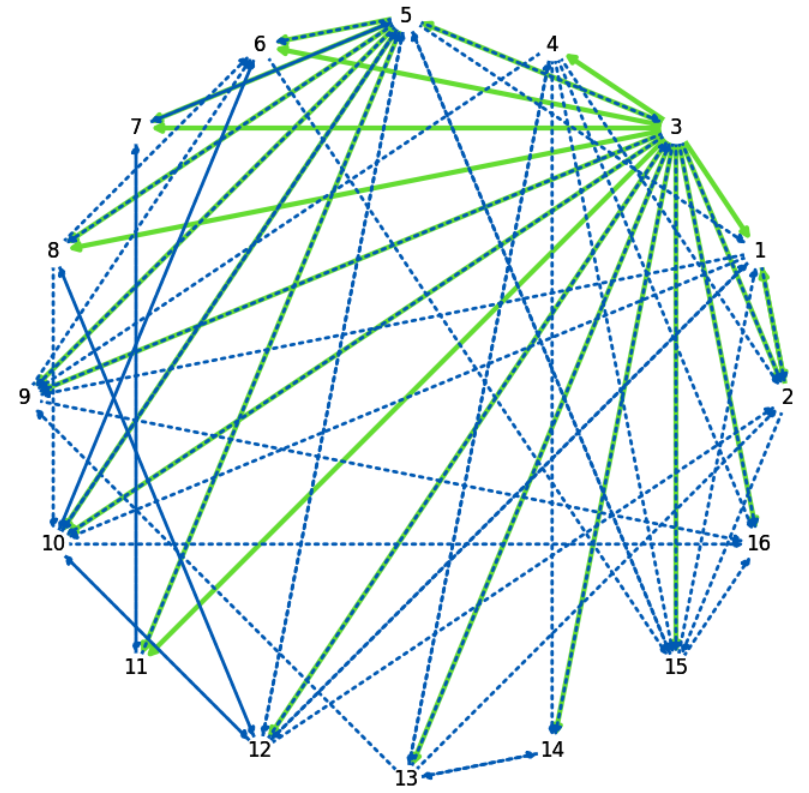}
    \caption{Example of an inferred GRN. Green : ground truth; Blue : predicted.}
\end{subfigure}
\caption{\textbf{Our model is competitive with state-of-the-art methods for GRN inference with orders of magnitude faster inference.} (a) We compare the ability of our model to predict the next states (dynamic accuracy) and the influence graph (structural accuracy) with that of other methods using a recent benchmark~\citep{puvsnik2022review} -- more details in Sec.~\ref{sec:benchmark}. (b) Average inference time of the various methods. (c) From the Boolean formulas predicted, one can construct an influence graph where each node represents a gene, and each arrow signals that one gene regulates another. }
\label{fig:bnet}
\end{figure*}
\paragraph{Benchmark}

We use the recent benchmark for GRN inference introduced by~\cite{puvsnik2022review}. This benchmark compares 5 methods for Boolean network inference on 30 Boolean networks inferred from biological data, with sizes ranging from 16 to 64 genes, and assesses both dynamical prediction (how well the model predicts the dynamics of the network) and structural prediction (how well the model predicts the Boolean functions compared to the ground truth). Structural prediction is the binary classification task of predicting whether variable $i$ influences variable $j$, and can be evaluated by many binary classification metrics; we report here the structural F1 and Youden's J statistic metrics, which are the most holistic, and defer other metrics to App.~\ref{app:grn}.

\paragraph{Method}

Our model predicts each component $f_i$ of the Boolean network independently, by taking as input the whole state of the network at times $[0\ldots t-1]$ and as output the state of the $i$th bit at times $[1\ldots t]$. Once each component has been predicted, we can build a causal influence graph, where an arrow connects node $i$ to node $j$ if $j$ appears in the update equation of $i$: an example is shown in Fig.~\ref{fig:bnet}c. Note that since the dynamics of the Boolean network tend to be slow, an easy way to get rather high dynamical accuracy would be to simply predict the trivial fixed point $f_i = x_i$. In concurrent approaches, the function set explored excludes this solution; in our case, we simply mask the $i$th bit from the input when predicting $f_i$.

\paragraph{Results}
We display the results of our model on the benchmark in Fig.~\ref{fig:bnet}a. Boolformer performs on par with the SOTA algorithms, GABNI~\citep{barman2018boolean} and MIBNI~\citep{barman2017novel}. A striking feature of our model is its inference speed, displayed in Fig.~\ref{fig:bnet}b: a few seconds, against up to an hour for concurrent approaches, which mainly rely on genetic programming. Note also that our model predicts an interpretable Boolean function, where the other SOTA methods (GABNI and MIBNI) only pick out the most important variables and the sign of their influence. 

The fast inference speed of the Boolformer suggests that it could be used in combination with genetic approaches, to further increase the quality of the results, at the cost of inference speed. In such an approach, the fast model can be used to replace the random mutations, instead sampling the mutations from Boolformer, which should allow for a much more efficient exploration of the search space \cite{romera-paredes_mathematical_2024}. Due to the autoregressive nature of Boolformer, it is not well-suited for this task; a BERT-like architecture \cite{devlin2019bertpretrainingdeepbidirectional} would be more appropriate, so we leave explorations of this approach for future work.

\section{Discussion}
\label{sec:discussion}

In this work, we have shown that Transformers can be used to strongly perform the symbolic regression of logical functions, opening up a new, more interpretable framework than classical machine learning to solve certain types of classification tasks. Their ability to infer GRNs several orders of magnitude faster than existing methods offers the promise of many other exciting applications in biology, where Boolean modelling plays a key role~\citep{hemedan2022boolean}. There are however several limitations in our current approach, which open directions for future work.
\vspace{-0.1cm}
\paragraph{Limited Number of Input Points.}
First, the number of input points is limited to a thousand during training, which limits our model's performance on high-dimensional functions (although the model does exhibit some length generalization abilities at inference, as shown in App.~\ref{app:length_gen}). Note that we did not consider linear attention mechanisms~\citep{choromanski2020rethinking,wang2020linformer} because on top of potentially degrading performance\footnote{We hypothesize that full attention span is particularly important in this specific task: the attention maps displayed in App.~\ref{app:attention} are visually dense and high-rank matrices.}, this would not fundamentally improve scalability, as the volume of input space grows exponentially with the input dimension.

\paragraph{Predefined Feature Sets.}
Our model is developed on binary input features. Although it is easy to binarize categorical and continuous features, this increases the input dimension significantly, and our model has a hard limit on the number of input features it can handle, which is set to a hundred in this work. This could be mitigated by considering an extension to q-ary input features, although this requires choosing a new list of associated universal operators.
\vspace{-0.1cm}
\paragraph{Vocabulary Limitations.}
The logical functions on which our model is trained do not include the XOR gate explicitly, limiting the compactness of the formulas it predicts. This limitation is due to our generation procedure that relies on expression simplification, which requires rewriting the XOR gate in terms of AND, OR and NOT. We leave it as future work to adapt the generation of simplified formulas containing XOR gates, as well as more general operators with more than two inputs as in~\cite{rosenberg2023explainable}.
\paragraph{Lack of Intermediate Results and Multi-Output.}
The simplicity of the formulas predicted is limited in two additional ways: our model only handles (i) single-output functions -- multi-output functions are predicted independently component-wise and (ii) gates with a fan-out of one. As a result, our model cannot reuse intermediary results for different outputs or for different computations within a single output\footnote{Consider the $D$-parity: one can build a formula with only $3(n-1)$ binary AND-OR gates by storing $D-1$ intermediary results: $a_1=XOR(x_1,x_2), a_2=XOR(a_1,x_3),\ldots, a_{n-1} = XOR(a_{D-2}, x_{D})$. Our model needs to recompute these intermediary values, leading to much larger formulas, e.g. 35 binary gates instead of 9 for the 4-parity as illustrated in App.~\ref{app:examples}.}. One could address this either by post-processing the generated formulas to identify repeated substructures, or by adapting the data generation process to support multi-output functions and cyclic graphs.

%\footnote{Note that although the fan-in is fixed to 2 during training, it is easy to transform the predictions to larger fan-in by merging ORs and ANDs together.}
Finally, this paper mainly focused on investigating concrete applications and benchmarks to motivate the potential and development of Boolformers. A natural direction is to investigate theoretically and practically how the control of the data generator can influence the model simplicity and its impact on the `generalization on the unseen'~\citep{abbe2023sgd} benchmarks. 

\paragraph{Acknowledgements}
The authors would like to thank Clément Hongler, Philippe Schwaller, Geemi Wellawatte, Enric Boix-Adsera, Alexander Mathis, François Charton as well as the anonymous reviewers for insightful discussions. We also thank Russ Webb, Samira Abnar and Omid Saremi for valuable thoughts and feedback on this work. SD acknowledges funding from the EPFL AI4science program when this work was carried out.

\clearpage
\printbibliography

%%%%%%%%%%%%%%%%%%%%%%%%%%%%%%%%%%%%%%%%%%%%%%%%%%%%%%%%%%%%%%%%%%%%
% APPENDIX
%%%%%%%%%%%%%%%%%%%%%%%%%%%%%%%%%%%%%%%%%%%%%%%%%%%%%%%%%%%%%%%%%%%%
\newpage
\appendix
\section{Details on data generation}
\label{app:generator}

\subsection{Formula generation}

To construct random unary-binary trees, we follow the steps below:
\begin{enumerate}[leftmargin=*,noitemsep]
    \item \textbf{Sample the input dimension} $\din$ of the function $f$ uniformly in $[1,\din_\text{max}]$ .
    \item \textbf{Sample the number of active variables} $\dactive$ uniformly in $[1, \min(\din, \dactive_\text{max})]$. $\dactive$ determines the number of variables which affect the output of $f$ (the number of active tree inputs); the other variables are inactive. Select a set of $\dactive$ variables from the original $\din$ variables uniformly at random.
    \item \textbf{Sample the number of binary operators} $B$ uniformly in $[\dactive-1,  B_\text{max}]$ then sample $B$ operators from \{AND, OR\} independently with equal probability.
    \item \textbf{Build a binary tree} with those $B$ nodes, using the sampling procedure of~\cite{lample2019deep}, designed to produce a diverse mix of deep and narrow versus shallow and wide trees.
    \item \textbf{Negate some of the nodes} of the tree by adding  NOT gates independently with  probability $p_\text{NOT}=\nicefrac{1}{2}$.
    \item \textbf{Fill in the leaves}: for each of the $B+1$ leaves in the tree, sample independently and uniformly at random one of the variables from the set of active variables\footnote{The  $S$ variables are sampled without replacement in order for all the active variables to appear in the tree.}.
    \item \textbf{Simplify} the tree using Boolean algebraic rules, as described below. This greatly reduces the number of operators, and occasionally reduces the number of active variables.
\end{enumerate}

To maximize diversity, we sample large formulas (up to $B_\text{max}=500$ binary gates), which are then heavily pruned in the simplification step\footnote{The simplification leads to a non-uniform distribution of number of operators as discussed in App.~\ref{app:generator}.}. 

\subsection{Formula simplification}

The data generation procedure heavily relies on expression simplification. This is of utmost importance for four reasons:
\begin{itemize}
    \item It reduces the output expression length and hence memory usage as well as increasing speed
    \item It improves the supervision by reducing expressions to a more canonical form, easier to guess for the model
    \item It encourages the model to output the simplest formula, which is a desirable property.
\end{itemize}

We use the package $\texttt{boolean.py}$\footnote{\url{https://github.com/bastikr/Boolean.py}} for this, which is considerably faster than $\texttt{sympy}$ (the function $\texttt{simplify\_logic}$ of the latter has exponential complexity, and is hence only implemented for functions with less than 9 input variables).

Empirically, we found the following procedure to be optimal in terms of average length obtained after simplification:
\begin{enumerate}
    \item Preprocess the formula by applying basic logical equivalences: double negation elimination and De Morgan's laws.
    \item Parse the formula with $\texttt{boolean.py}$ and run the $\texttt{simplify()}$ method until it \emph{stabilizes} (sometimes, simplify more than once is necessary)
    \item Apply once again the first step
\end{enumerate}

\begin{figure}[h!]
    \centering
    \includegraphics[width=.6\linewidth]{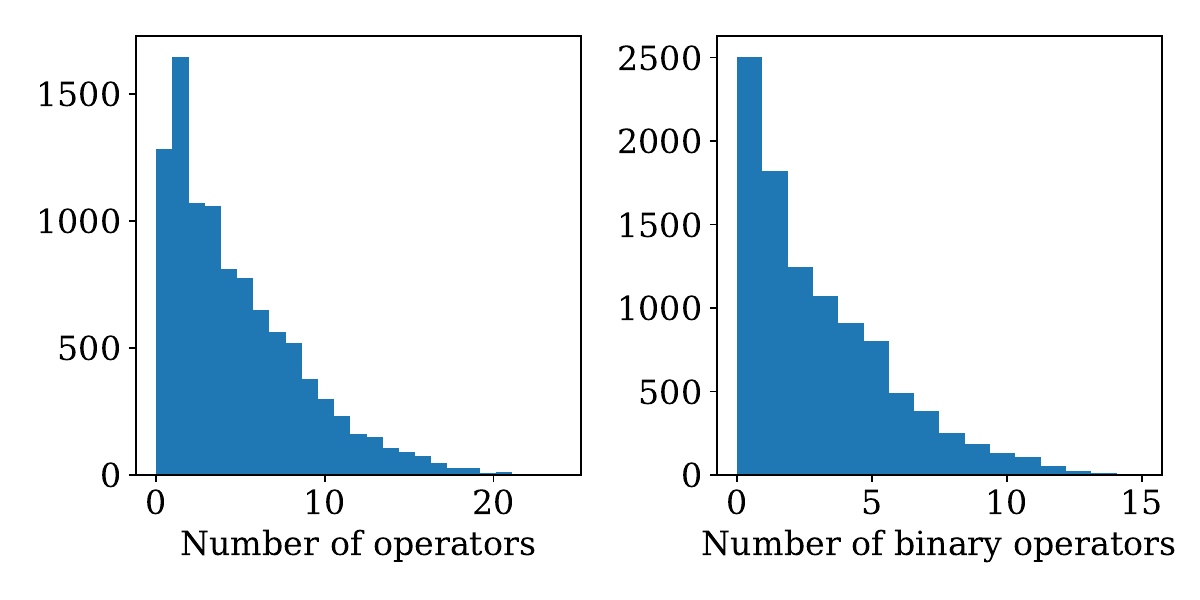}
    \caption{\textbf{Distribution of number of operators after expression simplification.} The initial number of binary operators is sampled uniformly in [1, 500]. The total number of examples is $10^4$.}
    \label{fig:histogram}
\end{figure}

Note that this procedure drastically reduces the number of operators and renders the final distribution highly nonuniform, as shown in Fig.~\ref{fig:histogram}.

\section{Does Boolformer memorize?}
\label{app:memorization}

One natural question is whether our model simply performs memorization on the training set. Indeed, the number of possible functions of $\din$ variables is finite, and equal to $2^{2^\din}$. 

Let us first assume naively that our generator is uniform in the space of boolean functions. Since $2^{2^4}\simeq 6\times 10^4$ (which is smaller than the number of examples seen during training) and $2^{2^5}\simeq 5.10^9$ (which is much larger), one could conclude that for $D\leq 4$, all functions are memorized, whereas for $D>4$, only a small subset of all possible functions are seen, hence memorization cannot occur.

However, the effective number of unique functions seen during training is actually smaller because our generator of random functions is nonuniform in the space of boolean functions. In this case, for which value of $\din$ does memorization become impossible? To investigate this question, for each $\din<\din_\text{max}$, we sample $\min\left(2^{2^D},100\right)$ unique functions from our random generator, and count how many times their exact truth table is encountered over an epoch (300,000 examples).

\begin{figure}[h!]
    \centering
    \includegraphics[width=.5\linewidth]{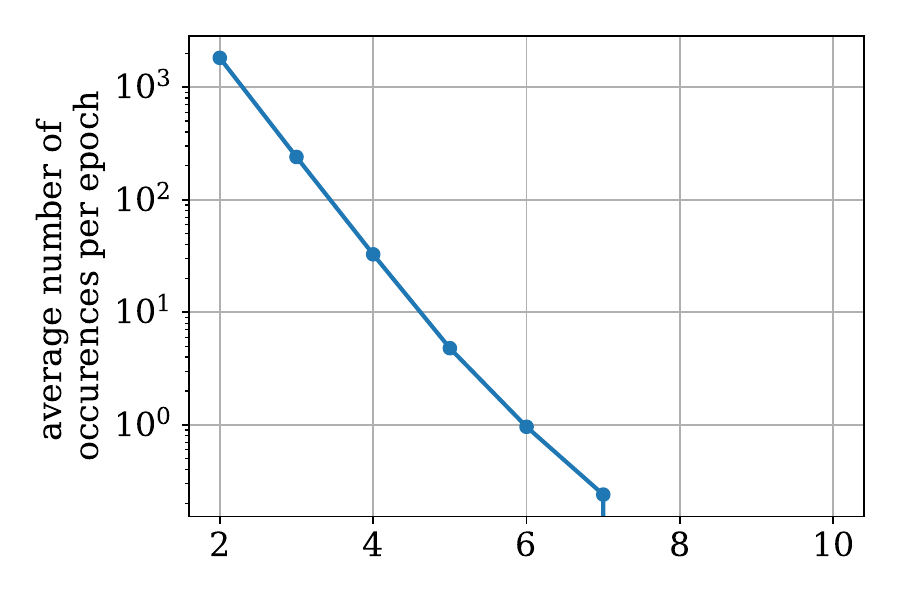}
    \caption{\textbf{Functions with 7 or more variables are typically never seen more than once during training.} We display the average number of times functions of various input dimensionalities are seen during an epoch (300,000 examples). For each point on the curve, the average is taken over $\min(2^{2^D}),100)$ unique functions.}
    \label{fig:redundancy}
\end{figure}

Results are displayed in Fig.~\ref{fig:redundancy}. As expected, the average number of occurences of each function decays exponentially fast, and falls to zero for $\din= 7$, meaning that each function is typically unique for $\din\geq7$. Hence, memorization cannot occur for $\din\geq7$. Yet, as shown in Fig.~\ref{fig:error-acc-noiseless}, our model achieves excellent accuracies even for functions of 10 variables, which excludes memorization as a possible explanation for the ability of our model to predict logical formulas accurately.

\section{Comparison to ESPRESSO in the noiseless setting}\label{app:espressocompare}
To compare the logic synthesis capability of Boolformer with ESPRESSO, we generate 3000 formulas with our generator. The number of input variables is distributed uniformly among [1,10], however, the final distribution is a bit different, as some of the inputs are sometimes redundant. We run these through Boolformer, which outputs valid simplifications (i.e. perfect recovery) on $96.5\%$ of the formulas in this set. We compare performance only on this subset of formulas, the remaining $3.5\%$ can be considered as 'failed to simplify' by Boolformer. 

\begin{figure}[h!]
    \centering
    \includegraphics[width=0.8\linewidth]{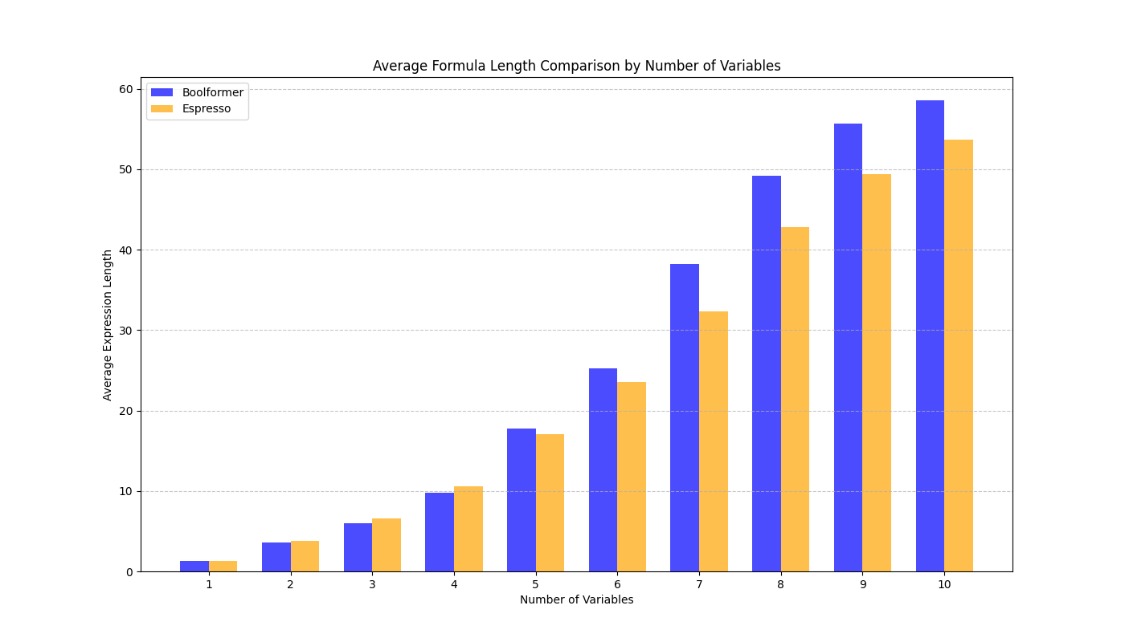}
    \caption{Histogram comparing the average length of output expression of Boolformer vs ESPRESSO. }
    \label{fig:avglength}
\end{figure}

\begin{figure}[h!]
    \centering
    \includegraphics[width=0.75\linewidth]{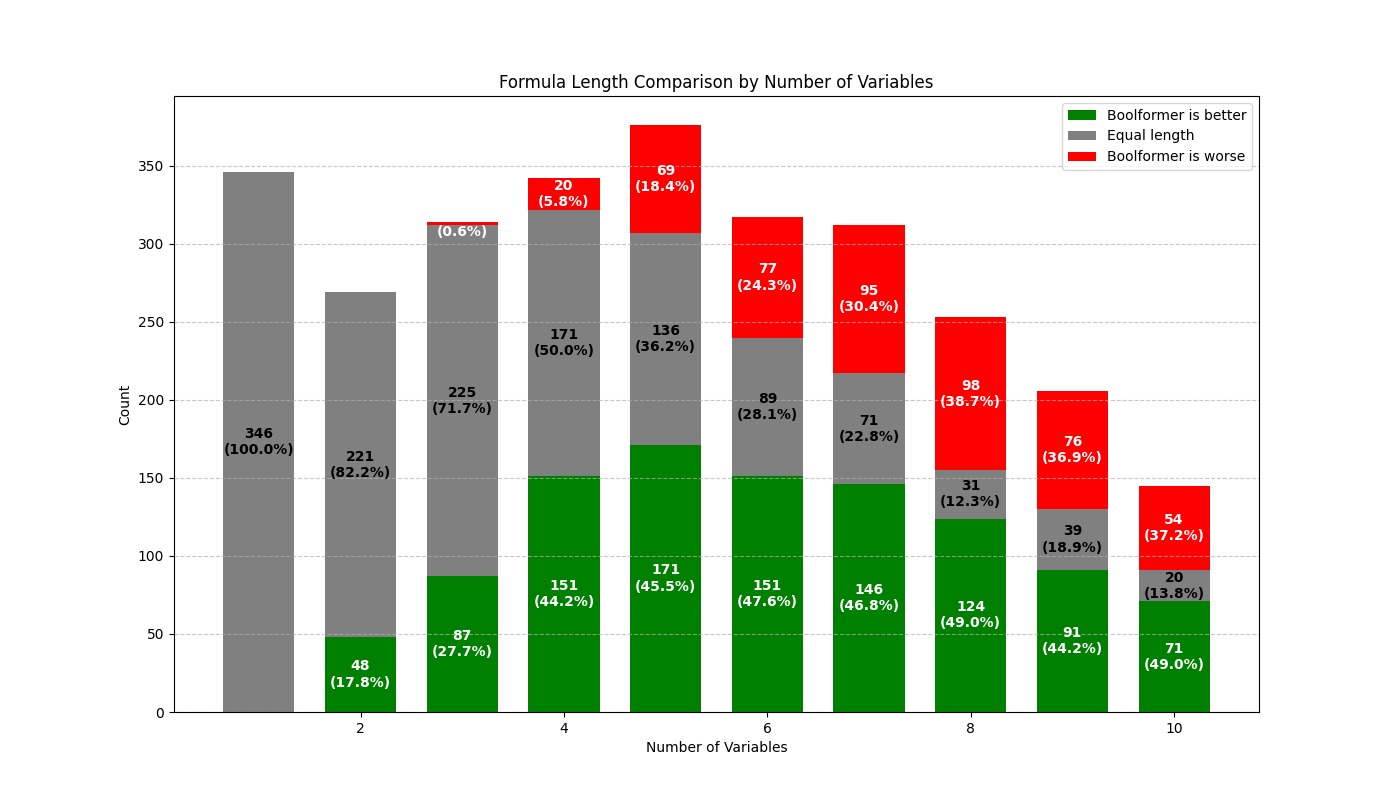}
    \caption{ \footnotesize Histogram displaying the head-to-head simplification performance of Boolformer vs ESPRESSO. The green portion is the percentage of formulas that were better simplified by Boolformer, the gray where both methods found equal length formulas, and the red where ESPRESSO did better. Formula lengths below 5 should be mostly disregarded, as  Boolformer is potentially able to ‘memorize’ the dataset at these lengths (see Fig \ref{fig:redundancy} ). However, we see that for longer formulas, when Boolformer succeeds, it is often shorter than ESPRESSO.}
    \label{fig:head-to-head}
\end{figure}

Fig. \ref{fig:avglength} displays the average length of the simplified formula for different numbers of active variables. Overall, Boolformer does worse on this front. One reason why the average length metric is not better for Boolformer is that when it fails to find a short simplification, it will sometimes try to correct with many tokens, generating a very long answer. Indeed, if we compare the 'head-to-head' count of which formula is shorter, we obtain Fig. \ref{fig:head-to-head}

Looking at Fig.\ref{fig:head-to-head}, Boolformer simplifications seem to be better than ESPRESSO a majority of the time, though not always. This conclusively shows that what Boolformer does is at least non-trivial: it does not consist of simple combinations (e.g. Disjunctive Normal Form, DNF) to craft a formula from its truth table. Additionally, its simplification capabilities extend beyond memorization, as they still produce good results for a number of variables above 7. That being said, the output space of Boolformer is bigger than ESPRESSO, as the former is allowed to output any valid Boolean formula using AND, OR, and NOT operations, while the latter has to output it in DNF. Nonetheless, it is the closest comparison we were able to find, as most logic synthesis programs are focused on multi-output functions, and mostly work with And-Inverter Graphs (AIGs), which are composed of an even more restrictive set of operations (only AND and NOT).

Finally, we can also compare the inference time per formula, which gives $0.06s$ for Boolformer and $3.16s$ for ESPRESSO. Here, the big advantage is mainly due to the fact that ESPRESSO runs on CPU, while Boolformer can leverage GPU, so it is not a particularly illuminating comparison. Still, maybe in future work one could focus on the noiseless setting, using a more reasonable representation as input (such as AIG, which is widely used in modern logic synthesis tools such as ABC \cite{berkeley-abcabc_2025}), and test to see whether Boolformer-like pipelines can advance the state of the art.
\section{Length generalization}
\label{app:length_gen}

In this section we examine the ability of our model to length generalize. In this setting, there are two types of generalization one can define: generalization in terms of the number of inputs $N$, or in terms of the number of active variables $S$\footnote{Note that our model cannot generalize to a problem of higher dimensionality $D$ than seen during training, as its vocabulary only contains the names of variables ranging from $x_1$ to $x_{\din_\text{max}}$.}. We examine length generalization in the  noisy setup (see Sec.~\ref{sec:inputs}), because in the noiseless setup, the model already has access to all the truth table (increasing $N$ does not bring any extra information), and all the variables are active (we cannot increase $S$ as it is already equal to $D$).

\subsection{Number of inputs}
\begin{figure}[h!]
    \centering
    \includegraphics[width=\linewidth]{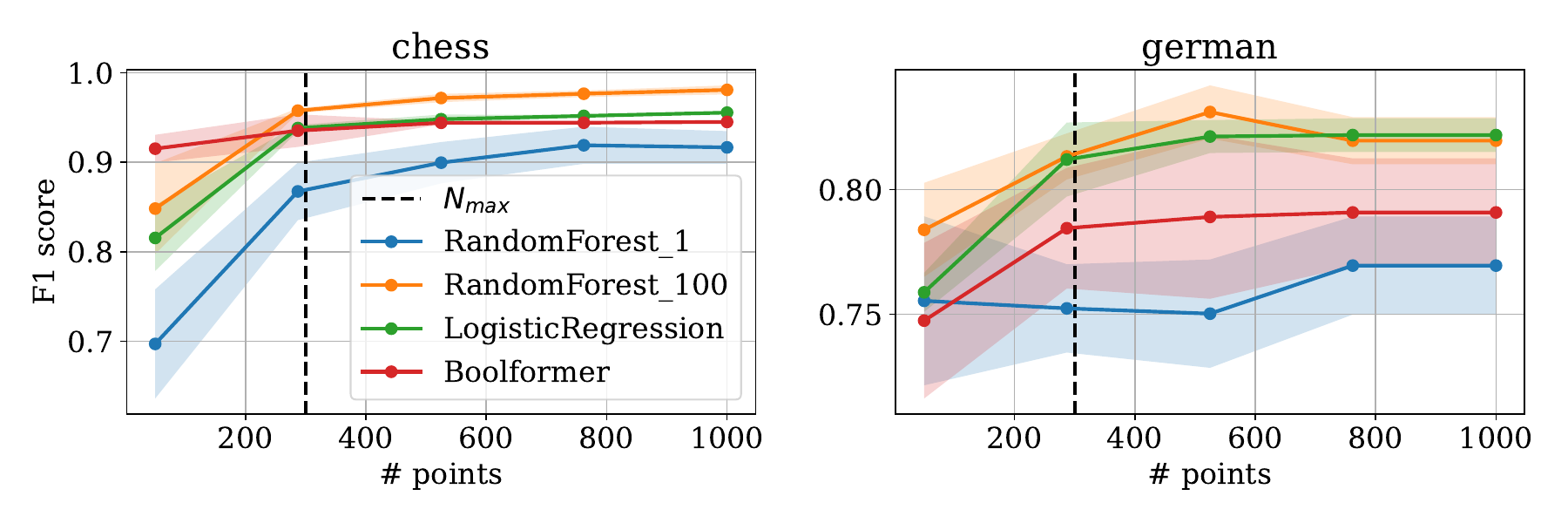}
    \caption{\textbf{Our model can length generalize in terms of sequence length.} We test a model trained with $N_\text{max}=300$ on the $\texttt{chess}$ and $\texttt{german}$ datasets of PMLB. Results are averaged over 10 random samplings of the input points, with the shaded areas depicting the standard deviation.}
    \label{fig:pmlb-num-points}
\end{figure}

Since the input points fed to the model are permutation invariant, our model does not use any positional embeddings. Hence, not only can our model handle $N>N_\text{max}$, but performance often continues to improve beyond $N_\text{max}$, as we show for two datasets extracted from PMLB~\citep{Olson2017PMLB} in Fig.~\ref{fig:pmlb-num-points}. 

\subsection{Number of variables}

To assess whether our model can infer functions which contain more active variables than seen during training, we evaluated a model trained on functions with up to 6 active variables on functions with 7 or more active variables. We provided the model with the truth table of two very simple functions: the OR and AND of the first $S\geq 7$ variables. We observe that the model succeeds for $S=7$, but fails for $S\geq 8$, where it only includes the first 7 variables in the OR / AND. Hence, the model can length generalize to a small extent in terms of number of active variables, but less easily than in terms of number of inputs. We hypothesize that proper length generalization could be achieved by "priming", i.e. adding even a small number of "long" examples, as performed in~\cite{jelassi2023length}.

\section{Formulas predicted for logical circuits}
\label{app:examples}

In Figs.~\ref{fig:arithmetic-formulas} and \ref{fig:logical_formulas}, we show examples of some common arithmetic and logical formulas predicted by our model in the noiseless regime, with a beam size of 100. In all cases, we increase the dimensionality of the problem until the failure point of Boolformer.

\begin{figure}[htb]
    \begin{subfigure}[b]{\linewidth}
        \small
        \centering
        \begin{forest}
        [$y_0$ [$\mathrm{or}$ [$\mathrm{and}$ [$x_0$ ][$x_2$ ]][$\mathrm{and}$ [$x_1$ ][$x_3$ ][$\mathrm{or}$ [$x_0$ ][$x_2$ ]]]]]
        \end{forest}
        \centering
        \begin{forest}
        [$y_1$ [$\mathrm{and}$ [$\mathrm{or}$ [$x_1$ ][$x_3$ ]][$\mathrm{not}$ [$\mathrm{and}$ [$x_1$ ][$x_3$ ]]]]]
        \end{forest}
        \centering
        \begin{forest}
        [$y_2$ [$\mathrm{and}$ [$\mathrm{or}$ [$x_0$ ][$\mathrm{not}$ [$\mathrm{and}$ [$x_1$ ][$x_2$ ][$x_3$ ]]]][$\mathrm{or}$ [$\mathrm{and}$ [$x_0$ ][$\mathrm{not}$ [$\mathrm{or}$ [$x_2$ ][$\mathrm{and}$ [$x_1$ ][$x_3$ ]]]]][$\mathrm{and}$ [$\mathrm{not}$ [$x_0$ ]][$x_2$ ]][$\mathrm{and}$ [$x_1$ ][$x_3$ ][$\mathrm{or}$ [$\mathrm{not}$ [$x_0$ ]][$x_2$ ]]]]]]
        \end{forest}
    \caption{Addition of two 2-bit numbers: $y_0y_1y_2=(x_0x_1)+(x_2x_3)$. All formulas are correct.}
    \end{subfigure}
    \begin{subfigure}[b]{\linewidth}
        \tiny
        \centering
        \begin{forest}
        [$y_0$ [$\mathrm{and}$ [$\mathrm{or}$ [$x_0$ ][$x_3$ ]][$\mathrm{or}$ [$\mathrm{and}$ [$x_0$ ][$x_3$ ]][$\mathrm{and}$ [$x_1$ ][$x_4$ ]][$\mathrm{and}$ [$x_2$ ][$x_5$ ][$\mathrm{or}$ [$x_1$ ][$x_4$ ]]]]]]
        \end{forest}
        \centering
        \begin{forest}
        [$y_1$ [$\mathrm{and}$ [$\mathrm{or}$ [$x_2$ ][$x_5$ ]][$\mathrm{not}$ [$\mathrm{and}$ [$x_2$ ][$x_5$ ]]]]]
        \end{forest}
        \centering
        \begin{forest}
        [$y_2$ [$\mathrm{not}$ [$\mathrm{or}$ [$\mathrm{and}$ [$x_1$ ][$x_4$ ][$\mathrm{not}$ [$\mathrm{and}$ [$x_2$ ][$x_5$ ]]]][$\mathrm{not}$ [$\mathrm{or}$ [$x_1$ ][$x_4$ ][$\mathrm{and}$ [$x_2$ ][$x_5$ ]]]][$\mathrm{and}$ [$x_2$ ][$x_5$ ][$\mathrm{or}$ [$x_1$ ][$x_4$ ]][$\mathrm{not}$ [$\mathrm{and}$ [$x_1$ ][$x_4$ ]]]]]]]
        \end{forest}
        \centering
        \begin{forest}
        [$y_3$ [$\mathrm{or}$ [$\mathrm{and}$ [$x_0$ ][$\mathrm{not}$ [$\mathrm{or}$ [$x_3$ ][$\mathrm{and}$ [$x_1$ ][$x_4$ ]][$\mathrm{and}$ [$x_2$ ][$x_5$ ]]]]][$\mathrm{and}$ [$\mathrm{not}$ [$x_0$ ]][$x_3$ ][$\mathrm{not}$ [$\mathrm{or}$ [$\mathrm{and}$ [$x_1$ ][$x_4$ ]][$\mathrm{and}$ [$x_2$ ][$x_4$ ][$x_5$ ]]]]][$\mathrm{and}$ [$\mathrm{or}$ [$x_0$ ][$\mathrm{not}$ [$x_3$ ]]][$\mathrm{or}$ [$\mathrm{not}$ [$x_0$ ]][$x_3$ ]][$\mathrm{or}$ [$x_1$ ][$\mathrm{and}$ [$x_2$ ][$x_5$ ]]][$\mathrm{or}$ [$x_4$ ][$\mathrm{and}$ [$x_1$ ][$x_2$ ][$x_5$ ]]]]]]
            \end{forest}
    \caption{Addition of two 3-bit numbers: $y_0y_1y_2y_3=(x_0x_1x_2)+(x_3x_4x_5)$. All formulas are correct, except $y_3$ which gets an error of 3\%.}
    \end{subfigure}
\end{figure}%
\begin{figure}[ht]\ContinuedFloat  
    \begin{subfigure}[b]{\linewidth}
        \small
        \centering
        \begin{forest}
        [$y_0$ [$\mathrm{and}$ [$x_0$ ][$x_1$ ][$x_2$ ][$x_3$ ]]]
        \end{forest}
        \centering
        \begin{forest}
        [$y_1$ [$\mathrm{and}$ [$x_1$ ][$x_3$ ]]]
        \end{forest}
        \centering
        \begin{forest}
        [$y_2$ [$\mathrm{and}$ [$\mathrm{not}$ [$\mathrm{and}$ [$x_0$ ][$x_1$ ][$x_2$ ][$x_3$ ]]][$\mathrm{or}$ [$\mathrm{and}$ [$x_0$ ][$x_3$ ]][$\mathrm{and}$ [$x_1$ ][$x_2$ ]]]]]
        \end{forest}
        \centering
        \begin{forest}
        [$y_3$ [$\mathrm{and}$ [$x_0$ ][$x_2$ ][$\mathrm{not}$ [$\mathrm{and}$ [$x_1$ ][$x_3$ ]]]]]
        \end{forest}
    \caption{Multiplication of two 2-bit numbers: $y_0y_1y_2y_3=(x_0x_1)\times (x_2x_3)$. All formulas are correct.}
    \end{subfigure}
    \begin{subfigure}[b]{\linewidth}
        \tiny
        \centering
        \centering
        \begin{forest}
        [$y_0$ [$\mathrm{and}$ [$x_0$ ][$x_3$ ][$\mathrm{or}$ [$\mathrm{and}$ [$x_1$ ][$x_4$ ]][$\mathrm{and}$ [$x_2$ ][$x_5$ ][$\mathrm{or}$ [$x_1$ ][$x_4$ ]]]]]]
        \end{forest}
        \centering
        \begin{forest}
        [$y_1$ [$\mathrm{and}$ [$x_2$ ][$x_5$ ]]]
        \end{forest}
        \centering
        \begin{forest}
        [$y_2$ [$\mathrm{and}$ [$\mathrm{not}$ [$\mathrm{and}$ [$x_1$ ][$x_2$ ][$x_4$ ][$x_5$ ]]][$\mathrm{or}$ [$\mathrm{and}$ [$x_1$ ][$x_5$ ]][$\mathrm{and}$ [$x_2$ ][$x_4$ ]]]]]
        \end{forest}
        \centering
        \begin{forest}
        [$y_3$ [$\mathrm{not}$ [$\mathrm{or}$ [$\mathrm{and}$ [$x_0$ ][$x_5$ ][$\mathrm{or}$ [$x_2$ ][$\mathrm{and}$ [$x_1$ ][$x_4$ ]]][$\mathrm{or}$ [$\mathrm{not}$ [$x_2$ ]][$x_3$ ]]][$\mathrm{and}$ [$x_1$ ][$x_2$ ][$x_3$ ][$x_4$ ][$\mathrm{not}$ [$x_5$ ]]][$\mathrm{and}$ [$\mathrm{not}$ [$\mathrm{and}$ [$x_0$ ][$x_5$ ]]][$\mathrm{or}$ [$\mathrm{not}$ [$x_1$ ]][$\mathrm{not}$ [$x_4$ ]][$\mathrm{and}$ [$x_2$ ][$x_5$ ]]][$\mathrm{not}$ [$\mathrm{and}$ [$x_2$ ][$x_3$ ]]]]]]]
        \end{forest}
        \centering
        \begin{forest}
        [$y_4$ [$\mathrm{and}$ [$\mathrm{or}$ [$\mathrm{not}$ [$x_0$ ]][$\mathrm{not}$ [$x_1$ ]][$x_2$ ][$\mathrm{not}$ [$x_3$ ]][$\mathrm{not}$ [$x_4$ ]][$x_5$ ]][$\mathrm{not}$ [$\mathrm{and}$ [$x_0$ ][$x_1$ ][$x_2$ ][$x_5$ ]]][$\mathrm{or}$ [$\mathrm{and}$ [$x_0$ ][$x_4$ ]][$\mathrm{and}$ [$x_1$ ][$x_2$ ][$\mathrm{not}$ [$x_3$ ]][$x_4$ ][$x_5$ ]][$\mathrm{and}$ [$x_3$ ][$\mathrm{or}$ [$x_1$ ][$\mathrm{and}$ [$x_0$ ][$x_2$ ][$x_5$ ]]][$\mathrm{not}$ [$\mathrm{and}$ [$x_2$ ][$x_4$ ]]]]]]]
        \end{forest}
        \centering
        \begin{forest}
        [$y_5$ [$\mathrm{not}$ [$\mathrm{and}$ [$\mathrm{or}$ [$\mathrm{not}$ [$x_0$ ]][$\mathrm{not}$ [$x_3$ ]][$\mathrm{and}$ [$x_1$ ][$x_4$ ]][$\mathrm{and}$ [$x_2$ ][$x_5$ ][$\mathrm{or}$ [$x_1$ ][$x_4$ ]]]][$\mathrm{or}$ [$\mathrm{not}$ [$x_1$ ]][$\mathrm{not}$ [$x_4$ ]][$\mathrm{and}$ [$x_0$ ][$\mathrm{not}$ [$x_5$ ]]][$\mathrm{not}$ [$\mathrm{or}$ [$x_0$ ][$x_3$ ]]][$\mathrm{and}$ [$\mathrm{not}$ [$x_2$ ]][$x_3$ ]]]]]]
        \end{forest}
    \caption{Multiplication of two 3-bit numbers: $y_0y_1y_2y_3y_4y_5=(x_0x_1x_2)\times (x_3x_4x_5)$. All formulas are correct, except $y_4$ which gets an error of 5\%.}
    \end{subfigure}
%\end{figure}%
% \begin{figure}[ht]\ContinuedFloat 
%     \begin{subfigure}[b]{\linewidth}
%         \small
%         \centering
%         \begin{forest}
%         [$y_0$ [$\mathrm{not}$ [$\mathrm{or}$ [$\mathrm{and}$ [$\mathrm{not}$ [$x_0$ ]][$x_4$ ]][$\mathrm{and}$ [$\mathrm{or}$ [$\mathrm{not}$ [$x_0$ ]][$x_4$ ]][$\mathrm{or}$ [$\mathrm{and}$ [$\mathrm{not}$ [$x_1$ ]][$x_5$ ]][$\mathrm{and}$ [$\mathrm{or}$ [$\mathrm{not}$ [$x_1$ ]][$x_5$ ]][$\mathrm{or}$ [$\mathrm{not}$ [$x_2$ ]][$x_6$ ]][$\mathrm{or}$ [$\mathrm{not}$ [$x_3$ ]][$x_7$ ][$\mathrm{and}$ [$\mathrm{not}$ [$x_2$ ]][$x_6$ ]]]]]]]]]
%         \end{forest}
%     \caption{Comparison of two 4-bit numbers: $y_0=(x_0x_1x_2x_3)>(x_4x_5x_6x_7)$. This formula is correct.}
%     \end{subfigure}
%     \begin{subfigure}[b]{\linewidth}
%         \small
%         \centering
%         \begin{forest}
%         [$y_0$ [$\mathrm{and}$ [$\mathrm{or}$ [$x_0$ ][$\mathrm{not}$ [$x_5$ ]]][$\mathrm{or}$ [$\mathrm{and}$ [$x_0$ ][$\mathrm{not}$ [$x_5$ ]]][$\mathrm{and}$ [$x_1$ ][$\mathrm{not}$ [$x_6$ ]]][$\mathrm{and}$ [$\mathrm{or}$ [$x_1$ ][$\mathrm{not}$ [$x_6$ ]]][$\mathrm{or}$ [$x_2$ ][$\mathrm{not}$ [$x_7$ ]]][$\mathrm{or}$ [$\mathrm{and}$ [$x_2$ ][$\mathrm{not}$ [$x_7$ ]]][$\mathrm{and}$ [$x_3$ ][$\mathrm{not}$ [$x_8$ ]]][$\mathrm{and}$ [$x_4$ ][$\mathrm{not}$ [$x_9$ ]][$\mathrm{or}$ [$x_3$ ][$\mathrm{not}$ [$x_8$ ]]]]]]]]]
%         \end{forest}
%     \caption{Comparison of two 5-bit numbers: $y_0=(x_0x_1x_2x_3x_4)>(x_5x_6x_7x_8x_9)$. This formula is correct.}
%     \end{subfigure}
    \caption{\textbf{Some arithmetic formulas predicted by our model.}}
    \label{fig:arithmetic-formulas}
\end{figure}

\begin{figure}[htb]
    \centering
    \begin{subfigure}[b]{\linewidth}
        \tiny
        \centering
        \begin{forest}
            [$y_0$ [$\mathrm{or}$ [$\mathrm{and}$ [$x_0$ ][$x_1$ ][$\mathrm{not}$ [$x_2$ ]][$\mathrm{not}$ [$x_3$ ]]][$\mathrm{and}$ [$x_0$ ][$\mathrm{not}$ [$x_1$ ]][$x_2$ ][$\mathrm{not}$ [$x_3$ ]]][$\mathrm{and}$ [$\mathrm{not}$ [$x_0$ ]][$x_1$ ][$x_2$ ][$\mathrm{not}$ [$x_3$ ]]][$\mathrm{and}$ [$\mathrm{not}$ [$x_0$ ]][$\mathrm{not}$ [$x_1$ ]][$x_2$ ][$x_3$ ]][$\mathrm{and}$ [$x_1$ ][$x_3$ ][$\mathrm{or}$ [$x_0$ ][$\mathrm{not}$ [$x_2$ ]]][$\mathrm{or}$ [$\mathrm{not}$ [$x_0$ ]][$x_2$ ]]][$\mathrm{and}$ [$\mathrm{not}$ [$x_1$ ]][$\mathrm{not}$ [$x_2$ ]][$\mathrm{or}$ [$x_0$ ][$\mathrm{not}$ [$x_3$ ]]][$\mathrm{or}$ [$\mathrm{not}$ [$x_0$ ]][$x_3$ ]]]]]
            \end{forest}
        \caption{4-parity: 0\% error.}
        \end{subfigure}
        \begin{subfigure}[b]{\linewidth}
            \tiny
            \centering
            \begin{forest}
            [$y_0$ [$\mathrm{or}$ [$\mathrm{and}$ [$x_0$ ][$x_1$ ][$x_2$ ][$x_3$ ][$\mathrm{not}$ [$x_4$ ]]][$\mathrm{and}$ [$x_0$ ][$\mathrm{not}$ [$x_2$ ]][$x_3$ ][$x_4$ ]][$\mathrm{and}$ [$\mathrm{not}$ [$x_0$ ]][$x_1$ ][$x_2$ ][$\mathrm{not}$ [$\mathrm{or}$ [$x_3$ ][$x_4$ ]]]][$\mathrm{and}$ [$\mathrm{not}$ [$x_0$ ]][$x_2$ ][$x_3$ ][$x_4$ ]][$\mathrm{and}$ [$\mathrm{not}$ [$x_1$ ]][$\mathrm{not}$ [$x_2$ ]][$\mathrm{not}$ [$x_3$ ]][$\mathrm{or}$ [$x_0$ ][$\mathrm{not}$ [$x_4$ ]]][$\mathrm{or}$ [$\mathrm{not}$ [$x_0$ ]][$x_4$ ]]][$\mathrm{and}$ [$x_2$ ][$\mathrm{not}$ [$x_3$ ]][$\mathrm{not}$ [$x_4$ ]][$\mathrm{or}$ [$x_0$ ][$x_1$ ]]][$\mathrm{and}$ [$\mathrm{not}$ [$x_2$ ]][$x_3$ ][$x_4$ ][$\mathrm{or}$ [$x_0$ ][$\mathrm{not}$ [$x_1$ ]]]][$\mathrm{and}$ [$\mathrm{not}$ [$x_2$ ]][$\mathrm{not}$ [$x_3$ ]][$\mathrm{not}$ [$x_4$ ]][$\mathrm{or}$ [$x_0$ ][$\mathrm{not}$ [$x_1$ ]]][$\mathrm{or}$ [$\mathrm{not}$ [$x_0$ ]][$x_1$ ]]]]]
            \end{forest}
            \caption{5-parity: 28\% error.}
            \end{subfigure}
\end{figure}%
\begin{figure}[ht]\ContinuedFloat  
    \centering
% \end{figure}%
% \begin{figure}[ht]\ContinuedFloat  
    \begin{subfigure}[b]{\linewidth}
        \centering
        \small
        \begin{forest}
            [$y_0$ [$\mathrm{and}$ [$\mathrm{or}$ [$x_0$ ][$\mathrm{and}$ [$x_1$ ][$x_2$ ]][$\mathrm{and}$ [$x_3$ ][$x_4$ ]]][$\mathrm{or}$ [$x_1$ ][$x_2$ ][$\mathrm{and}$ [$x_0$ ][$x_3$ ][$x_4$ ]]][$\mathrm{or}$ [$x_3$ ][$x_4$ ][$\mathrm{and}$ [$x_0$ ][$x_1$ ][$x_2$ ]]]]]
            \end{forest}
        \caption{5-majority: 0\% error.}
        \end{subfigure}
    \begin{subfigure}[b]{\linewidth}
        \centering
        \small
        \begin{forest}
            [$y_0$ [$\mathrm{and}$ [$\mathrm{or}$ [$x_0$ ][$x_3$ ][$x_4$ ][$\mathrm{and}$ [$x_1$ ][$x_2$ ]]][$\mathrm{or}$ [$x_1$ ][$x_2$ ][$x_4$ ][$\mathrm{and}$ [$x_0$ ][$x_3$ ][$x_5$ ]]][$\mathrm{or}$ [$x_1$ ][$x_3$ ][$x_5$ ][$\mathrm{and}$ [$x_0$ ][$x_2$ ]]][$\mathrm{or}$ [$x_2$ ][$x_3$ ][$x_5$ ][$\mathrm{and}$ [$x_0$ ][$x_1$ ][$x_4$ ]]][$\mathrm{or}$ [$x_4$ ][$x_5$ ][$\mathrm{and}$ [$x_0$ ][$x_1$ ]][$\mathrm{and}$ [$x_2$ ][$x_3$ ][$\mathrm{or}$ [$x_0$ ][$x_1$ ]]]]]]
        \end{forest}
            \caption{6-majority: 6\% error.}
            \end{subfigure}
    \caption{\textbf{Some logical functions predicted by our model.} }
    \label{fig:logical_formulas}
\end{figure}

\section{Formulas predicted for PMLB datasets}
\label{app:pmlb}

In Fig.~\ref{fig:pmlb-formulas-app}, we report a few examples of boolean formulas predicted for the PMLB datasets in Fig.~\ref{fig:pmlb}. In each case, we also report the F1 scores of logistic regression and random forests with 100 estimators.

\begin{figure}

    \begin{subfigure}[b]{.48\linewidth}
        \centering
        \begin{forest}
        [$\mathrm{not}$ [$\mathrm{and}$ [$\mathrm{not}$ [$\substack{\text{A20}}$ ]][$\mathrm{or}$ [$\substack{\text{A32}}$ ][$\substack{\text{A09}}$ ][$\mathrm{not}$ [$\mathrm{or}$ [$\substack{\text{A20}}$ ][$\mathrm{not}$ [$\substack{\text{A31}}$ ]][$\substack{\text{A34}}$ ]]]]]]
        \end{forest}
        \caption{chess. F1: 0.947. LogReg: 0.958. RF: 0.987.}
        \end{subfigure}
        \begin{subfigure}[b]{.48\linewidth}
        \centering
        \begin{forest}
        [$\mathrm{not}$ [$\mathrm{or}$ [$\substack{\text{surgery=0}}$ ][$\mathrm{not}$ [$\substack{\text{surgery=1}}$ ]][$\mathrm{not}$ [$\substack{\text{outcome=0}}$ ]][$\mathrm{and}$ [$\mathrm{or}$ [$\substack{\text{surgery=0}}$ ][$\mathrm{not}$ [$\substack{\text{surgery=1}}$ ]]][$\mathrm{or}$ [$\substack{\text{surgery=0}}$ ][$\substack{\text{outcome=0}}$ ]]]]]
        \end{forest}
        \caption{horse colic. F1: 0.900. LogReg: 0.822. RF: 0.861.}
        \end{subfigure}
        \begin{subfigure}[b]{\linewidth}
        \centering
        \begin{forest}
        [$\mathrm{not}$ [$\mathrm{or}$ [$\substack{\text{pension=2}}$ ][$\mathrm{not}$ [$\mathrm{or}$ [$\substack{\text{contribution to} \\ \text{dental plan=1}}$ ][$\substack{\text{pension=0}}$ ][$\mathrm{not}$ [$\mathrm{or}$ [$\mathrm{not}$ [$\substack{\text{duration=2}}$ ]][$\mathrm{and}$ [$\mathrm{not}$ [$\substack{\text{contribution to} \\ \text{dental plan=1}}$ ]][$\substack{\text{pension=0}}$ ]]]]]]]]
        \end{forest}
        \caption{labor. F1: 0.960. LogReg: 1.000. RF: 1.000.}
        \end{subfigure}

    \end{figure}
    \begin{figure}[ht]\ContinuedFloat

        \begin{subfigure}[b]{.48\linewidth}
            \centering
            \begin{forest}
            [$\mathrm{or}$ [$\substack{\text{Jacket color=2}}$ ][$\mathrm{and}$ [$\substack{\text{Head shape=0}}$ ][$\substack{\text{Body shape=0}}$ ]][$\mathrm{and}$ [$\substack{\text{Head shape=2}}$ ][$\substack{\text{Body shape=2}}$ ]]]
            \end{forest}
            \caption{monk1. F1: 0.915. LogReg: 0.732. RF: 1.000.}
        \end{subfigure}
        \begin{subfigure}[b]{.48\linewidth}
        \centering
        \begin{forest}
        [$\mathrm{not}$ [$\mathrm{or}$ [$\substack{\text{Jacket color=0}}$ ][$\mathrm{and}$ [$\substack{\text{Body shape=0}}$ ][$\mathrm{not}$ [$\mathrm{and}$ [$\substack{\text{Holding=2}}$ ][$\substack{\text{Jacket color=1}}$ ]]]]]]
        \end{forest}
        \caption{monk3. F1: 1.000. LogReg: 0.985. RF: 0.993.}
        \end{subfigure}
        \begin{subfigure}[b]{.48\linewidth}
        \centering
        \begin{forest}
        [$\mathrm{or}$ [$\substack{\text{F11}}$ ][$\substack{\text{F13}}$ ][$\substack{\text{F16}}$ ][$\substack{\text{F20}}$ ][$\substack{\text{F22}}$ ][$\mathrm{and}$ [$\substack{\text{F10}}$ ][$\mathrm{or}$ [$\substack{\text{F11}}$ ][$\substack{\text{F16}}$ ][$\mathrm{not}$ [$\substack{\text{F20}}$ ]]]]]
        \end{forest}
        \caption{spect. F1: 0.919. LogReg: 0.930. RF: 0.909.}
        \end{subfigure}        
        \begin{subfigure}[b]{.48\linewidth}
        \centering
        \begin{forest}
        [$\mathrm{not}$ [$\mathrm{or}$ [$\substack{\text{physician fee} \\ \text{freeze=0}}$ ][$\mathrm{and}$ [$\mathrm{not}$ [$\substack{\text{physician fee} \\ \text{freeze=2}}$ ]][$\mathrm{or}$ [$\mathrm{and}$ [$\substack{\text{physician fee} \\ \text{freeze=2}}$ ][$\mathrm{not}$ [$\substack{\text{physician fee} \\ \text{freeze=0}}$ ]]][$\mathrm{not}$ [$\mathrm{or}$ [$\substack{\text{physician fee} \\ \text{freeze=2}}$ ][$\substack{\text{education spending=2}}$ ]]]]]]]
        \end{forest}
        \caption{vote. F1: 0.971. LogReg: 0.974. RF: 0.974.}
        \end{subfigure}       

    \caption{\textbf{Some logical formulas predicted by our noisy model for some binary classification PMLB datasets.} In each case, we report the name of the dataset and the F1 score of the Boolformer, logistic regression and random forest in the caption.}
    \label{fig:pmlb-formulas-app}
\end{figure}

\clearpage

\section{Additional results on gene regulatory network inference}
\label{app:grn}

In this section, we give a very brief overview of the field of GRN inference and present additional results using our Boolformer.

\subsection{A brief overview of GRNs}

Inferring the behavior of GRNs is a central problem in computational biology, which consists in deciphering the activation or inhibition of one gene by another gene from a set of noisy observations.  This task is very challenging due to the low signal-to-noise ratios recorded in biological systems, and the difficulty to obtain temporal ordering and ground truth networks. 

GRN algorithms can infer relationships between the genes based on static observations~\cite{singh2015blars,haury2012tigress,huynh2010inferring}, or on input time-series recordings~\cite{adabor2019restricted,huynh2018dyngenie3}, and can either infer correlational relationships, i.e. undirected graphs, or causal relationships, i.e. directed graphs -- the latter being more useful, but harder to obtain. 

We focus on methods which model the dynamics of GRNs via Boolean networks: REVEAL~\citep{liang1998reveal}, Best-Fit~\citep{lahdesmaki2003learning}, MIBNI~\citep{barman2017novel}, GABNI~\citep{barman2018boolean} and ATEN~\citep{shi2020aten}. We evaluate our approach on the recent benchmark from~\cite{puvsnik2022review}.

\subsection{Additional results}

\begin{figure}[tb]
        \centering
        \includegraphics[width=\linewidth]{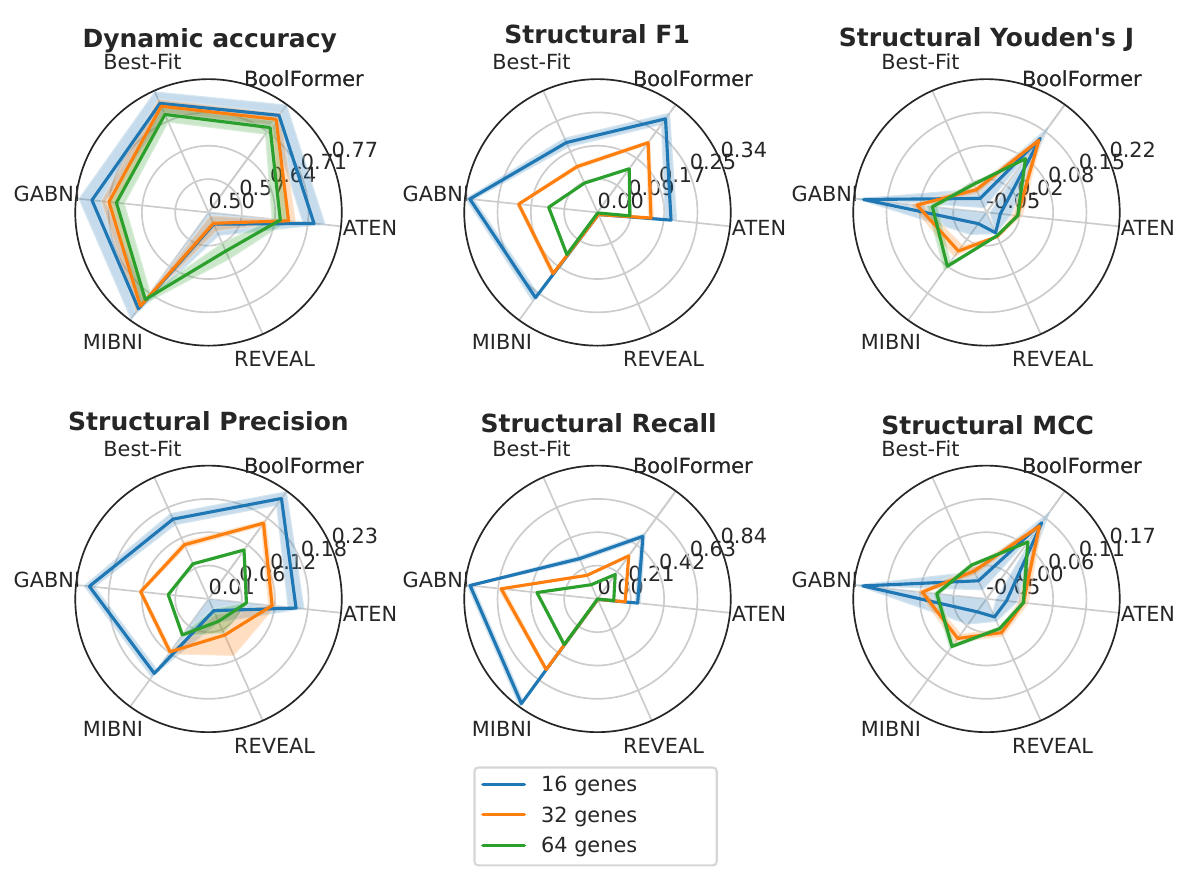}
    \caption{\textbf{Binary classification metrics used in the gene regulatory network benchmark.} The competitors and metrics are taken from the recent benchmark of~\cite{puvsnik2022review}, and described in Sec.~\ref{sec:benchmark}.}
    \label{fig:bnet-full}
    \end{figure}

The benchmark studied in the main text assesses both dynamical prediction (how well the model predicts the dynamics of the network) and structural prediction (how well the model predicts the Boolean functions compared to the ground truth). Structural prediction is framed as the binary classification task of predicting whether variable $i$ influences variable $j$, and can hence be evaluated by several binary classification metrics, defined below\footnote{The authors of the benchmark consider the two latter to be the best metrics to give a comprehensive view on the classifier performance for this task.}:
\begin{align*}
\text{Acc}=\frac{\text{TP}+\text{TN}}{\text{TP}+\text{TN}+\text{FP}+\text{FN}}&,\quad 
\text{Pre}=\frac{\text{TP}}{\text{TP}+\text{FP}},\quad
\text{Rec}=\frac{\text{TP}}{\text{TP}+\text{FN}},\quad
\text{F1}=2\frac{\text{Pre}\cdot \text{Rec}}{\text{Pre}+\text{Rec}},\\
\text{MCC}=&\frac{\text{TP}\cdot \text{TN}-\text{FP}\cdot \text{FN}}{\sqrt{(\text{TP}+\text{FP})(\text{TP}+\text{FN})(\text{TN}+\text{FP})(\text{TN}+\text{FN})}}, \text{BM} =  \frac{\text{TP}}{\text{TP}+\text{FN}}+\frac{\text{TN}}{\text{TN}+\text{FP}}-1
\end{align*}
Where TP, TN are True Positive and True Negative, while FP FN are False Positive and False Negative.
We report these metrics in Fig.~\ref{fig:bnet-full}.

\clearpage

\section{Exploring the beam candidates}
    \label{app:beam}
    
    In this section, we explore the beam candidates produced by the Boolformer. In Fig.~\ref{fig:beam}, we show the 8 top-ranked candidates when predicting a simple logic function, the 2-comparator. We see that all candidates perfectly match the ground truth, but have different structure. 
    
    \begin{figure}
        \small
        \centering
        \begin{forest}
        [$\mathrm{or}$ [$\mathrm{and}$ [$x_0$ ][$\mathrm{not}$ [$x_2$ ]]][$\mathrm{and}$ [$x_1$ ][$\mathrm{not}$ [$x_3$ ]][$\mathrm{or}$ [$x_0$ ][$\mathrm{not}$ [$x_2$ ]]]]]
        \end{forest}
        \centering \hfill
        \begin{forest}
        [$\mathrm{and}$ [$\mathrm{or}$ [$x_0$ ][$\mathrm{not}$ [$x_2$ ]]][$\mathrm{or}$ [$\mathrm{and}$ [$x_0$ ][$\mathrm{not}$ [$x_2$ ]]][$\mathrm{and}$ [$x_1$ ][$\mathrm{not}$ [$x_3$ ]]]]]
        \end{forest}
        \centering
        \begin{forest}
        [$\mathrm{not}$ [$\mathrm{and}$ [$\mathrm{or}$ [$\mathrm{not}$ [$x_0$ ]][$x_2$ ]][$\mathrm{or}$ [$\mathrm{not}$ [$x_1$ ]][$x_3$ ][$\mathrm{and}$ [$\mathrm{not}$ [$x_0$ ]][$x_2$ ]]]]]
        \end{forest}
        \centering \hfill
        \begin{forest}
        [$\mathrm{not}$ [$\mathrm{or}$ [$\mathrm{and}$ [$\mathrm{not}$ [$x_0$ ]][$x_2$ ]][$\mathrm{and}$ [$\mathrm{or}$ [$\mathrm{not}$ [$x_0$ ]][$x_2$ ]][$\mathrm{or}$ [$\mathrm{not}$ [$x_1$ ]][$x_3$ ]]]]]
        \end{forest}
        \centering
        \begin{forest}
        [$\mathrm{or}$ [$\mathrm{and}$ [$x_0$ ][$x_1$ ][$\mathrm{not}$ [$x_3$ ]]][$\mathrm{and}$ [$\mathrm{not}$ [$x_2$ ]][$\mathrm{or}$ [$x_0$ ][$\mathrm{and}$ [$x_1$ ][$\mathrm{not}$ [$x_3$ ]]]]]]
        \end{forest}
        \centering \hfill
        \begin{forest}
        [$\mathrm{or}$ [$\mathrm{and}$ [$x_0$ ][$\mathrm{or}$ [$\mathrm{not}$ [$x_2$ ]][$\mathrm{and}$ [$x_1$ ][$\mathrm{not}$ [$x_3$ ]]]]][$\mathrm{and}$ [$x_1$ ][$\mathrm{not}$ [$\mathrm{or}$ [$x_2$ ][$x_3$ ]]]]]
        \end{forest}
        \centering
        \begin{forest}
        [$\mathrm{or}$ [$\mathrm{and}$ [$x_0$ ][$\mathrm{not}$ [$x_2$ ]]][$\mathrm{and}$ [$x_1$ ][$\mathrm{not}$ [$x_3$ ]][$\mathrm{or}$ [$x_0$ ][$\mathrm{and}$ [$x_1$ ][$\mathrm{not}$ [$x_2$ ]]]]]]
        \end{forest}
        \centering \hfill
        \begin{forest}
        [$\mathrm{and}$ [$\mathrm{or}$ [$x_0$ ][$\mathrm{and}$ [$x_1$ ][$\mathrm{not}$ [$\mathrm{or}$ [$x_2$ ][$x_3$ ]]]]][$\mathrm{or}$ [$\mathrm{not}$ [$x_2$ ]][$\mathrm{and}$ [$x_1$ ][$\mathrm{not}$ [$x_3$ ]]]]]
        \end{forest}
    \caption{\textbf{Beam search reveals equivalent formulas}. We show the first 8 beam candidates for the 2-comparator, which given two 2-bit numbers $a=(x_0x_1)$ and $b = (x_2x_3)$, returns 1 if $a>b$, 0 otherwise. All candidates perfectly match the ground truth.}
    \label{fig:beam}
    \end{figure}

\section{Attention maps}
\label{app:attention}

In Fig.~\ref{fig:attention}, we show the attention maps produced by our model when presented three truth tables: (a) that of the 4-digit multiplier, (b) that of the 4-parity function and (c) a random truth table. Each panel corresponds to a different layer and head of the model.

Each attention map is an $N\times N$ matrix, where 
$N$ is the number of input points. The element 
$(i,j)$ represents the attention score between tokens 
$i$ and $j$, and is marked by the colormap, from blue (0) to yellow (1). Here the tokens are ordered from left to right by lexicographic order: 0000, 0001, 0010, ..., 1111. In this particular order, many interesting structures appear, especially for the first two functions which are non-random. For example, for the 4-parity function, the anti-diagonal attention map of (head 8, layer 7) indicates that the model compares antipodal points in the hypercube: (0000, 1111), (0011, 1100)...

As for the 4-digit multiplier, some attention heads have hadamard-like structure (e.g. heads 3,4,5 of layer 8), some have block-structured checkboard patterns (e.g. head 12 of layer 5), and many heads put most attention weight on the final input, 1111, which is more informative (e.g. head 15 of layer 3).

\begin{figure}[htb]
    \centering
    \begin{subfigure}[b]{\linewidth}
        \includegraphics[width=\linewidth]{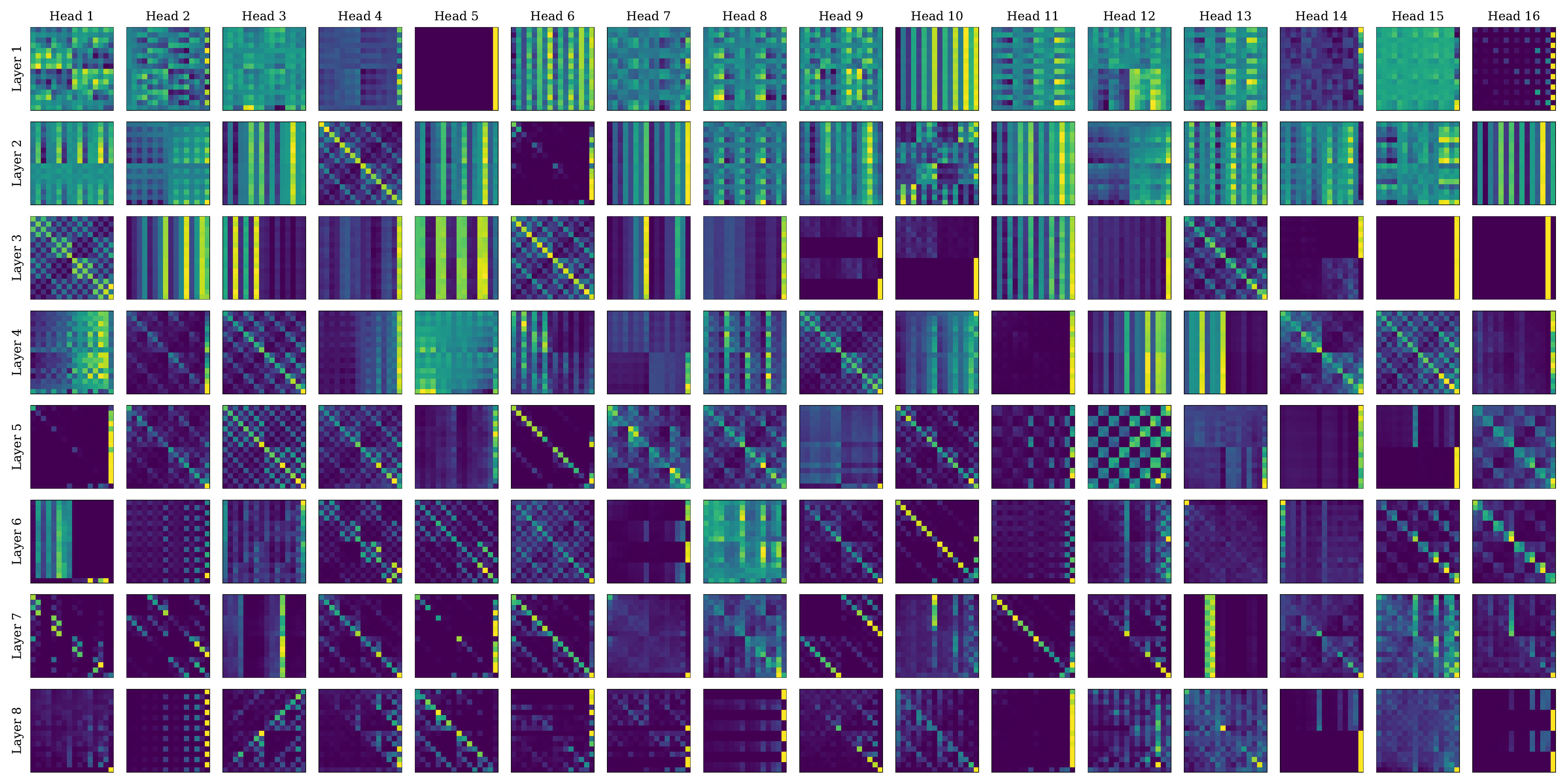}
        \caption{4-digit multiplier}
    \end{subfigure} 
    \begin{subfigure}[b]{\linewidth}
        \includegraphics[width=\linewidth]{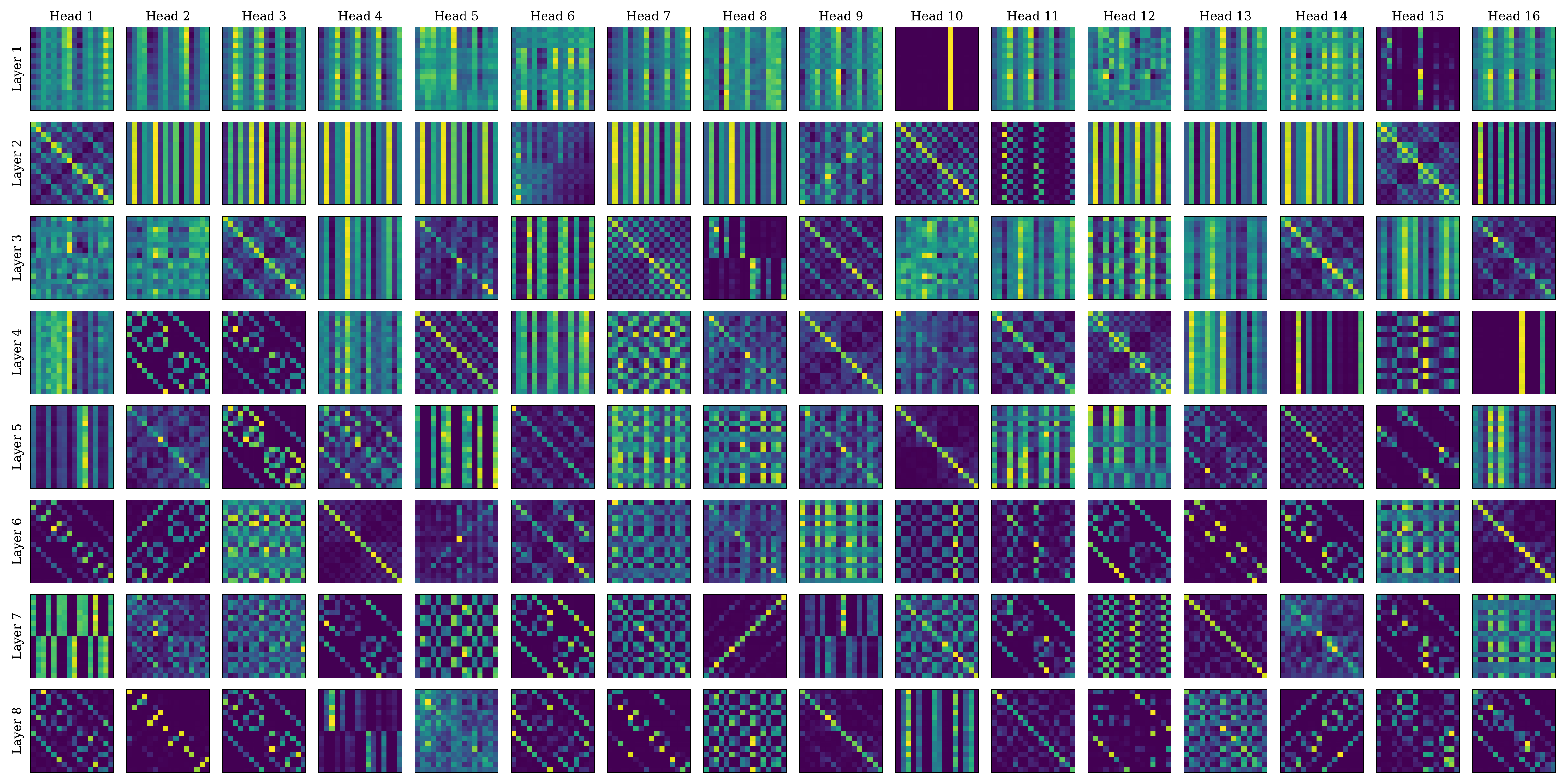}
        \caption{4-parity}
    \end{subfigure} 
\end{figure}%
\begin{figure}[ht]\ContinuedFloat 
    \begin{subfigure}[b]{\linewidth}
        \includegraphics[width=\linewidth]{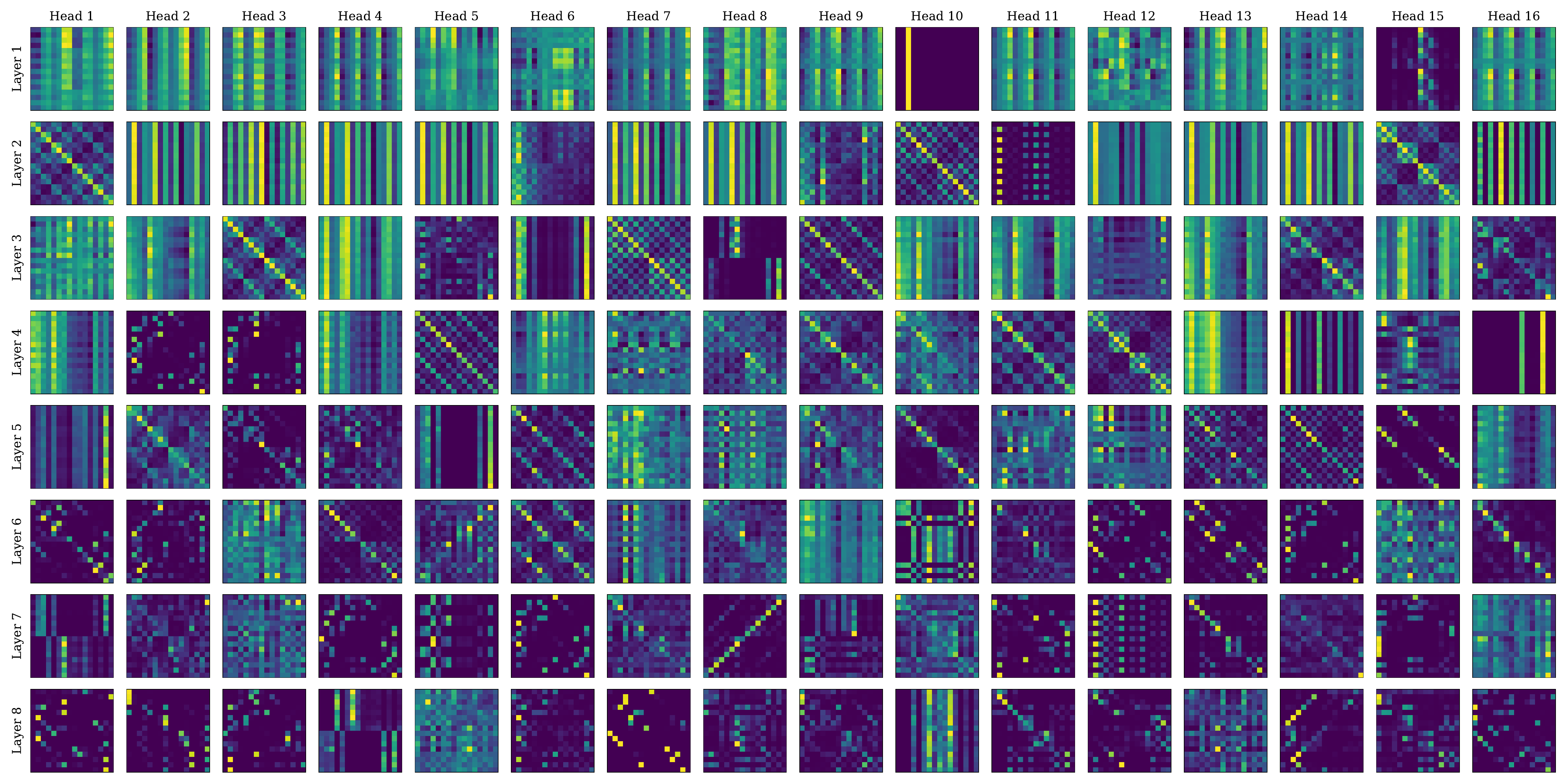}
        \caption{4d random data}
    \end{subfigure} 
    
    \caption{\textbf{The attention maps reveal intricate analysis.} See Sec.~\ref{app:attention} for more details on this figure.}
    \label{fig:attention}
\end{figure}

\end{document}